\documentclass[lettersize,journal]{IEEEtran}
\usepackage{amsmath,amsfonts}
\usepackage{algorithmic}
\usepackage{algorithm}
\usepackage{array}
\usepackage{textcomp}
\usepackage{stfloats}
\usepackage{url}
\usepackage{verbatim}
\usepackage{graphicx}
\usepackage{cite}
\usepackage{bbding}
\usepackage{booktabs}
\usepackage{float}
\usepackage{subfigure}
\usepackage{setspace}
\usepackage{soul}
\usepackage{color}

\def\BibTeX{{\rm B\kern-.05em{\sc i\kern-.025em b}\kern-.08em
    T\kern-.1667em\lower.7ex\hbox{E}\kern-.125emX}}
\usepackage{balance}
\begin{document}
\title{Strongly Augmented Contrastive Clustering}
\author{Xiaozhi Deng, Dong Huang,~\IEEEmembership{Member,~IEEE, }Ding-Hua Chen\\ 
        Chang-Dong Wang,~\IEEEmembership{Member,~IEEE, },
        Jian-Huang Lai,~\IEEEmembership{Senior Member,~IEEE, }
\IEEEcompsocitemizethanks{
	\IEEEcompsocthanksitem This project was supported by the NSFC (61976097, 61876193  \& 62076258) and the Natural Science Foundation of Guangdong Province (2021A1515012203). \textit{(Corresponding author: Dong Huang)}
	\IEEEcompsocthanksitem X. Deng, D. Huang, and D.-H. Chen are with the College of Mathematics and Informatics, South China Agricultural University, Guangzhou, China. \protect\\
	E-mail: dengxiaozhi45@gmail.com, huangdonghere@gmail.com, \protect\\
	dhchen@stu.scau.edu.cn.
	\IEEEcompsocthanksitem C.-D. Wang and J.-H. Lai are with the School of Computer Science and Engineering,
	Sun Yat-sen University, Guangzhou, China, and also with Guangdong Key Laboratory of Information Security Technology, Guangzhou, China, and also with Key Laboratory of Machine Intelligence and Advanced Computing, Ministry of Education, China.\protect\\
	E-mail: changdongwang@hotmail.com, stsljh@mail.sysu.edu.cn.}
}

\maketitle

\begin{abstract}
Deep clustering has attracted increasing attention in recent years due to its capability of joint representation learning and clustering via deep neural networks. In its latest developments, the contrastive learning has emerged as an effective technique to substantially enhance the deep clustering performance. However, the existing contrastive learning based deep clustering algorithms mostly focus on some carefully-designed augmentations (often with limited transformations to preserve the structure), referred to as weak augmentations, but cannot go beyond the weak augmentations to explore the more opportunities in stronger augmentations (with more aggressive transformations or even severe distortions). In this paper, we present an end-to-end deep clustering approach termed \textbf{S}trongly \textbf{A}ugmented \textbf{C}ontrastive \textbf{C}lustering (SACC), which extends the conventional two-augmentation-view paradigm to multiple views and jointly leverages strong and weak augmentations for strengthened deep clustering. Particularly, we utilize a backbone network with triply-shared weights, where a strongly augmented view and two weakly augmented views are incorporated. Based on the representations produced by the backbone, the weak-weak view pair and the strong-weak view pairs are simultaneously exploited for the instance-level contrastive learning (via an instance projector) and the cluster-level contrastive learning (via a cluster projector), which, together with the backbone, can be jointly optimized in a purely unsupervised manner. Experimental results on five challenging image datasets have shown the superiority of our SACC approach over the state-of-the-art. The code is available at \url{https://github.com/dengxiaozhi/SACC}.
\end{abstract}


\section{Introduction}\label{sec1}

\IEEEPARstart{D}{ata} clustering is a fundamental task in unsupervised learning, which aims to group a set of data samples into different unlabeled clusters. The traditional clustering algorithms, such as $K$-means \cite{macqueen1967some}, agglomerative clustering (AC) \cite{gowda1978agglomerative}, and spectral clustering (SC) \cite{zelnik2004self}, typically rely on the hand-crafted data features, which lack the representation learning ability and may lead to poor clustering performance when dealing with some complex high-dimensional data, such as images and videos, where the proper features are not easy to be manually extracted.

With the rapid development of deep learning, the deep neural network has recently been adopted to learn proper representations for the clustering task on complex high-dimensional data. In the past few years, many clustering algorithms based on deep neural networks (referred to as deep clustering algorithms) have been devised \cite{hershey2016deep,guo2017improved,caron2018deep,ji2019invariant,wang22_tcsvt}.
As one of the earliest deep clustering works, Xie et al. \cite{xie2016unsupervised} proposed a deep embedded clustering (DEC) method to simultaneously perform feature representations learning and clustering in a deep neural network, where the distribution of soft labels and an auxiliary target distribution are constrained via a Kullback-Leibler (KL) divergence based loss. Guo et al. \cite{guo2017improved} developed an improved deep embedded clustering (IDEC) method by learning the feature representation and the cluster assignment with local structure preservation. Caron et al. \cite{caron2018deep} iteratively clustered the learned features via $K$-means and update the weights of the deep neural network by using the cluster assignment as soft labels. Ji et al. \cite{ji2019invariant} designed a deep clustering method termed invariant information clustering (IIC), which seeks to maximize the mutual information between the original image and the augmented one for more robust representation learning and clustering.

\begin{figure}[!t]
\centering
\includegraphics[width=0.47\textwidth]{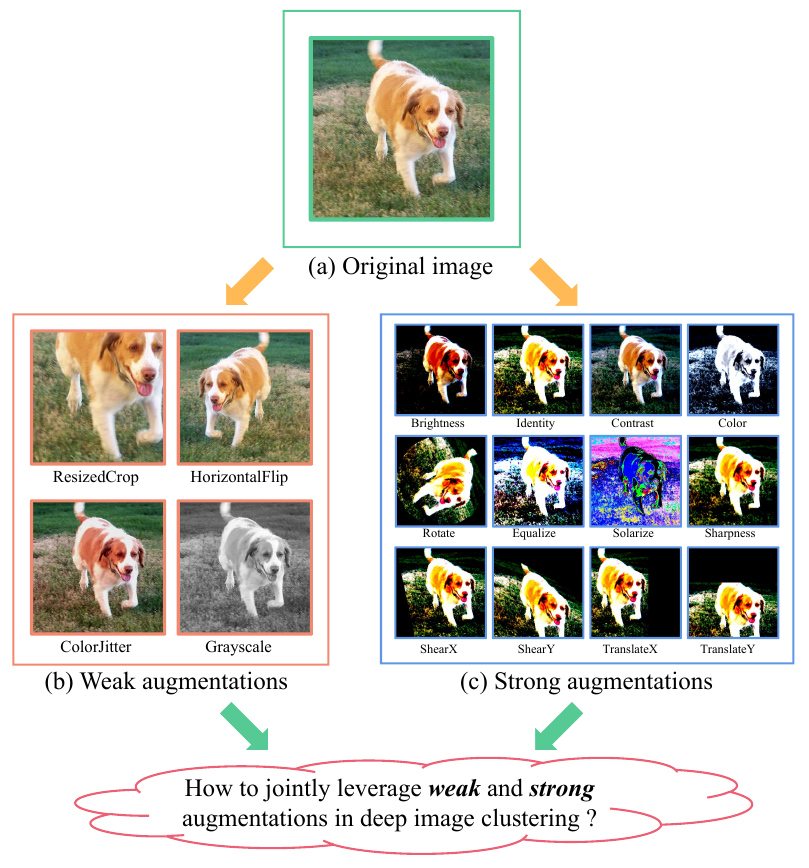}
\caption{The visualization of weak and strong augmentations. (a) Original image. (b) Weak augmentations. (c) Strong augmentations. }\vskip 0.02 in
\label{fig1}
\end{figure}

\begin{figure*}[!t]\vskip 0.05 in
\centering
\includegraphics[width=1\textwidth]{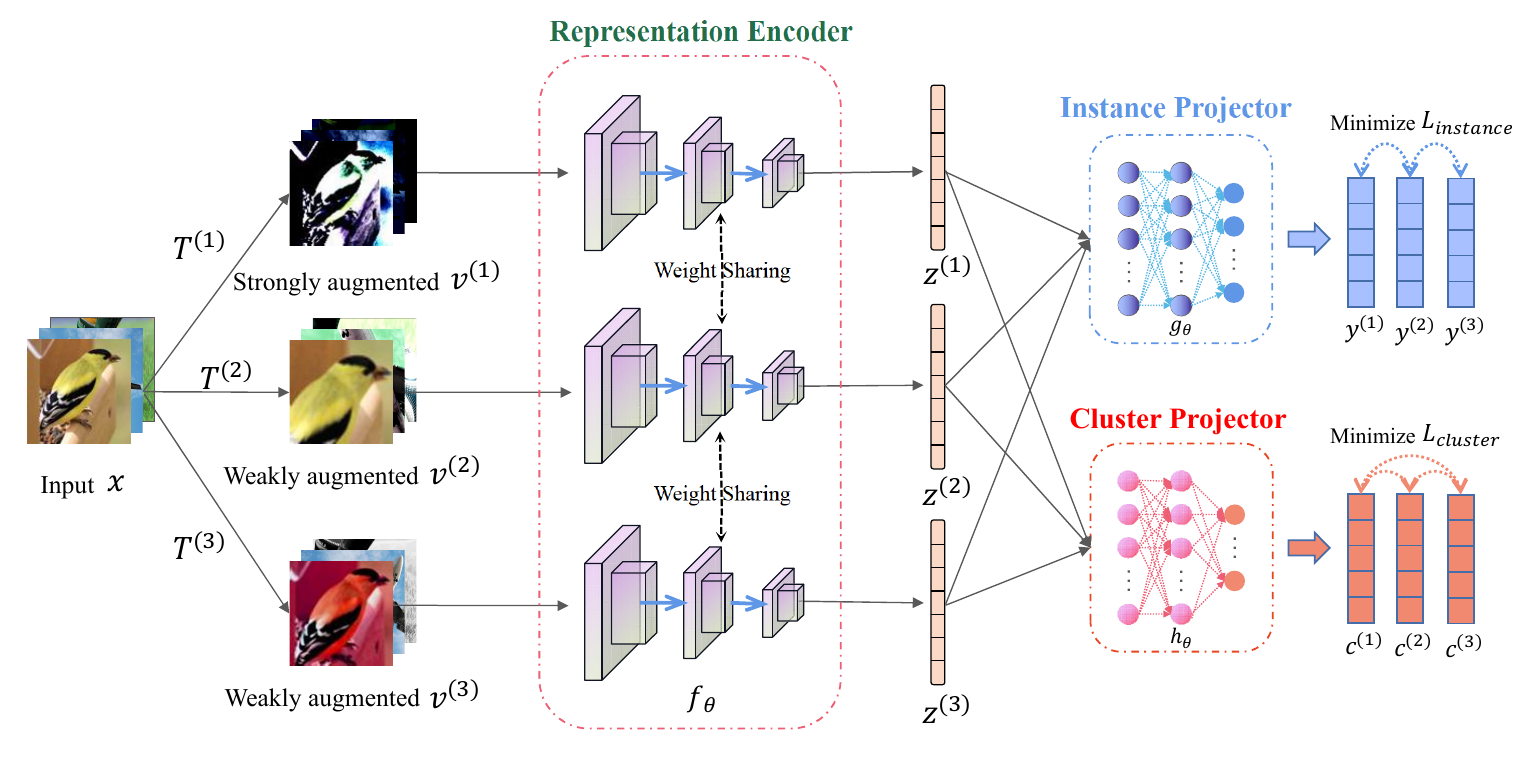}
\caption{Illustration of the proposed SACC framework. By simultaneously incorporating strong and weak augmentations, a backbone network with triply-shared weights is used to learn the representations of three augmentation views, which are then fed to two types of projectors for instance-level and cluster-level contrastive learning, respectively.}\vskip 0.1 in
\label{fig2}
\end{figure*}

Despite the considerable progress, these methods \cite{hershey2016deep,guo2017improved,caron2018deep,ji2019invariant} often perform the feature representation learning and clustering by considering the overall distributions (such as the distribution of soft labels or some other target distributions), which overlook the the sample-wise relationships and their contrastiveness. Recently, the contrastive learning has emerged as an effective technique for enhancing the deep clustering performance, which typically generates positive sample pairs and negative sample pairs via data augmentations, and aims to maximize the agreement between positive pairs and minimize the  agreement between negative pairs. For example, van Gansbeke et al. \cite{van2020scan} proposed a two-stage deep clustering method termed semantic clustering by adaptive nearest neighbors (SCAN), which first utilizes the contrastive learning to learn the discriminant features for finding the $K$ nearest neighbors, and then trains the network via a loss function that aims to pull each sample and its $K$ nearest neighbors closer. Dang et al. \cite{dang2021nearest} extended the SCAN method by matching both local and global nearest neighbors.
Li et al. \cite{li2021contrastive} devised a one-stage method termed contrastive clustering (CC), which jointly exploits instance-level and cluster-level contrastive learning in an end-to-end manner.

Though these contrastive learning based deep clustering methods \cite{van2020scan,dang2021nearest,li2021contrastive} have shown substantial improvements on some complex image datasets, yet there are still two limitations to most of them.
First, the previous deep clustering methods with contrastive learning tend to utilize weak augmentations (with limited transformations to preserve the structure) for original images, but mostly overlook the opportunities in stronger augmentations which may even be coupled with more aggressive transformations or distortions (as illustrated in Fig.~\ref{fig1}).
Second, they mostly design the network with two augmentation views (typically drawn from the same family of augmentations), but cannot go beyond two augmentation views to explore multiple views of augmentations (especially with varying degrees of transformations or distortions).
More recently, Wang and Qi \cite{wang2021contrastive} shown that the incorporation of stronger augmentations can enhance the feature representation learned by contrastive learning, which, however, is designed for the general-purpose contrastive learning but lacks the ability to achieve the representation learning and  clustering simultaneously. It remains an open problem how to simultaneously leverage the strong and weak augmentations while extending the conventional two-view network architecture to explore opportunities in multiple augmentation views in a unified deep clustering framework.

In light of this, this paper presents a novel end-to-end deep clustering approach termed \textbf{s}trongly \textbf{a}ugmented \textbf{c}ontrastive \textbf{c}lustering (SACC), which is able to jointly learn the feature representation and the cluster assignments with the strong and weak augmentations simultaneously leveraged in a network of multiple augmentation views (as illustrated in Fig.~\ref{fig2}).
In particular, our SACC approach utilizes a backbone network with triply-shared weights to produce the feature embeddings of a strongly augmented view and two weakly augmented views, upon which the weak-weak view pair and the strong-weak view pairs can be constructed for the instance-level and cluster-level contrastive learning. With the strong and weak augmentations as well as the instance-level and cluster-level contrastive learning jointly leveraged, the network training can thus be performed in a purely unsupervised manner and the deep clustering result is therefore obtained. Extensive experiments are conducted on five challenging image datasets, which demonstrate the superiority of our SACC approach over the state-of-the-art deep clustering approaches.

For clarity, the key contributions of this work are summarized as follows.
\begin{itemize}
  \item This paper for the first time, to the best of our knowledge, jointly leverages strong and weak augmentations for the task of unsupervised image clustering.
  \item A novel end-to-end deep clustering approach termed SACC is proposed, which utilizes three augmentations views for simultaneous instance-level and cluster-level contrastive learning..
  \item Extensive experimental results have confirmed that our SACC approach outperforms the state-of-the-art deep clustering approaches on several challenging image datasets.
\end{itemize}

The remainder of this paper is organized as follows. Section~\ref{sec2} reviews the related works on deep clustering. Section~\ref{sec3} describes the proposed SACC framework. Section~\ref{sec4} reports the experimental results. Finally, we conclude this paper in Section~\ref{sec5}.

\section{Related Work}\label{sec2}

Deep learning has proved to be an advantageous technique for unsupervised clustering of very complex data. Many deep clustering methods have been designed\cite{hershey2016deep,guo2017improved,caron2018deep,ji2019invariant,wang22_tcsvt,yang2017towards,ji2017deep,ghasedi2017deep,jiang2016variational,dilokthanakul2016deep,springenberg2015unsupervised,xiao21_tcsvt}, whose difference can often be reflected by their network losses, such as the reconstruction loss of autoencoder (AE), the variational loss of variational autoencoder (VAE) \cite{kingma2013auto}, the loss of generative adversarial network (GAN) \cite{goodfellow2014generative}, and some specific clustering losses \cite{yang2016joint,guo2019adaptive}.

The AE-based deep clustering methods generally optimize the networks by both the reconstruction loss and some clustering loss. The reconstruction loss measures the disagreement between the original input and the reconstruction. Yang et al. \cite{yang2017towards} presented the deep clustering network (DCN) method with the dimensionality reduction with the $K$-means clustering jointly modeled. Ji et al. \cite{ji2017deep} proposed an AE-based deep clustering method with a self-expressive layer for deep subspace clustering. Dizaji et al. \cite{ghasedi2017deep} developed the deep embedded regularized clustering (DEPICT) method based on AE embedding and relative entropy minimization.

The VAE-based deep clustering methods utilize the VAE to regularize the network training to avoid over-fitting by enforcing the latent space to follow some predefined distribution. Jiang et al. \cite{jiang2016variational} presented a variational deep embedding (VaDE) method that optimizes the VAE by maximizing the evidence lower bound. Dilokthanakul et al. \cite{dilokthanakul2016deep} proposed a Gaussian mixture variational autoencoder (GMVAE) method by incorporating a variational Bayes in its optimization objective.

The GAN-based deep clustering methods seek to train the network with a min-max adversarial game. Springenberg \cite{springenberg2015unsupervised} proposed a categorical generative adversarial network (CatGAN) method that jointly exploits GAN and regularized information maximization (RIM) to train the network. Chen et al. \cite{chen2016infogan} developed an information maximizing generative adversarial network (InfoGAN) method that aims to extract interpretable and disentangled features for deep clustering.

Different from the above three categories that usually combine the clustering loss with some network losses (such as the losses of AE, VAE, and GAN), another category of deep clustering methods aim to train the network with only the clustering loss. For example, Yang et al. \cite{yang2016joint} leveraged a convolutional neural network to learn representation feature and image clusters by a weighted triplet loss.
Xie et al. \cite{xie2016unsupervised} devised a deep embedding clustering (DEC) method that jointly optimizes deep embedding and clustering with a KL divergence based loss between the distribution of soft labels and an auxiliary target distribution. Guo et al. \cite{guo2019adaptive} developed an adaptive self-paced deep clustering with data augmentation (ASPC-DA) method that incorporates data augmentation and self-paced learning into deep clustering. 

Recently, the contrastive learning has become a popular topic \cite{chen2020simple}, and several attempts have been made to utilize the contrastive loss to improve the deep clustering performance \cite{van2020scan, dang2021nearest, li2021contrastive}. Typically, van Gansbeke et al. \cite{van2020scan} proposed a two-stage deep clustering method which adopts the contrastive learning as a pretext task to learn discriminant features and then exploits the $K$ nearest neighbors (via the learned features) in the second-stage network training. Dang et al. \cite{dang2021nearest} presented a nearest neighbor matching (NNM) method by considering not only the global nearest neighbors but also the local nearest neighbors. Li et al. \cite{li2021contrastive} performed contrastive learning at both instance-level and cluster-level and obtained the clustering result via a cluster projector.

\section{Proposed Framework}\label{sec3}

In this section, we describe the proposed SACC framework. Specifically, an overview the of the framework is given in Section~\ref{sec:framework_overview}. The weak and strong augmentations are introduced in Section~\ref{sec:weak_strong_aug}. The design of the network architecture is provided in Section~\ref{sec:architecture}. Finally, the implementation details are presented in Section~\ref{sec:implement_details}.

\subsection{Framework Overview}
\label{sec:framework_overview}

The overall framework of SACC is illustrated in Fig.~\ref{fig2}. In SACC, we utilize a backbone network with triply-shared weights, where the representations of three augmentation views (including one strong augmentation view and two weak augmentation views) are learned. Specifically, given a mini-batch of $N$ image samples, we performs one type of strong augmentation and two types of weak augmentations on each input image, denoted as $x_{i}$, which lead to $3\cdot N$ augmented samples denoted as $\{v_{1}^{1}, \ldots, v_{N}^{1}, v_{1}^{2}, \ldots, v_{N}^{2}, v_{1}^{3}, \ldots, v_{N}^{3}\}$,  with $N$ strongly augmented samples and $2\cdot N$ weakly augmented samples. The backbone network $f_{\theta}$ transforms each augmented sample $v_{i}^{j}$ to $z_{i}^{j}$, with $i \in[1, N]$ and $j \in\{1, 2, 3\}$, which will then be fed to the instance projector and the cluster projector. With the instance projector $g_{\theta}$ transforming  $z_{i}^{j}$ to $y_{i}^{j}$ and the cluster predictor $h_{\theta}$ transforming $z_{i}^{j}$ to $c_{i}^{j}$, two types of feature matrices are built (for this mini-batch of samples) via the two projectors, respectively. Thereafter, unsupervised network training can be performed by simultaneous optimizing the instance-level contrastive loss (in the row space of the feature matrix in the instance projector) and the cluster-level contrastive loss (in the column space of the feature matrix in the cluster projector).

\subsection{Augmentations: From Weak to Strong}
\label{sec:weak_strong_aug}

The contrastive learning has shown its promising ability in unsupervised representation learning \cite{chen2020simple}, and has been utilized in some recent deep clustering methods \cite{van2020scan,dang2021nearest,li2021contrastive}.
In previous deep clustering methods with contrastive learning, some weak augmentations (with limited transformations to preserve the image structure) are generally exploited to form the positive pairs. However, few of them have gone beyond the weak augmentations to utilize some stronger augmentations (with more aggressive transformations or even severe distortions).

In this paper, we have shown that the joint use of strong and weak augmentations can substantially strengthen the contrastive learning ability for representation learning and clustering in the deep clustering framework.
Specifically, in terms of the weak augmentations, we adopt a family of four often-used augmentations, namely, ResizedCrop, HorizontalFlip, ColorJitter, and Grayscale, to generate weakly augmented samples. For the two weak augmentation views in SACC, two augmentations are randomly chosen from the family of weak augmentations for each input image.
Besides the weak augmentations, we adopt a family of fourteen stronger transformations \cite{wang2021contrastive}, including  AutoContrast, Brightness, Color, Contrast, Equalize, Identity, Posterize, Rotate, Sharpness, ShearX/Y, Solarize and TranslateX/Y. Since the strong augmentations transform the original image more aggressively, it can provide some additional clues that do not exist in the weak augmentations for learning distinctive representations. By jointly modeling strong and weak augmentations, the proposed SACC framework is able to obtain more representative patterns and semantic information of images for learning clustering-friendly representations.

\subsection{Network Architecture}
\label{sec:architecture}

The network architecture of SACC consists of three modules, namely, the backbone network, the instance projector, and the cluster projector. An instance-level contrastive loss and an cluster-level contrastive loss are utilized in the instance projector and the cluster projector, respectively, which are jointly trained with both the weak and strong augmentation views. In the following, we will describe the three modules as well as the overall objective function in detail.

\subsubsection{Backbone Network with Triply-Shared Weights}

In the proposed framework, we utilize a backbone network with triply-shared weights (as shown in Fig.~\ref{fig2}). Specifically, three augmentation views, including a
strong augmentation view and two weak augmentation views share the backbone network $f_{\theta}$, through which three views of representations can be learned, denoted as $z_{i}^{1} = f(v_{i}^{1})$, $z_{i}^{2} = f(v_{i}^{2})$, and $z_{i}^{3} = f(v_{i}^{3})$. Then three feature representations are fed to each of the two projectors for the later instance-level and cluster-level contrastive learning. Note that we can adopt different network structures as the backbone. In this work, we adopt the widely-used ResNet-34 \cite{he2016deep} as our backbone network.

\subsubsection{Instance Projector with Weak-Strong Augmentations}

In the instance projector, a two-layer nonlinear multilayer perceptron (MLP), denoted as $g(\cdot)$, is used to transform $z_{i}^{j}$ to a lower-dimensional space,  that is, $y_{i}^{j} = g(z_{i}^{j})$, where $y_{i}^{j}$ is interpreted as the instance representation, with $i \in[1, N]$ and $j \in\{1, 2, 3\}$.

As there are three augmentation views, including two weak augmentation views and a strong augmentation view, we use a weak-weak pair and a strong-strong pair to form the positive pairs and the negative pairs. Specifically, for each input image, its two weakly augmented samples form a positive pair, and its strongly augmented sample and its first weakly augmented sample form another positive pair. In the meantime, negative pairs are formed between the augmented samples from different input images.

With the positive pairs and negative pairs defined, the instance-level contrastive loss is utilized to maximize the agreement of positive pairs while increasing the distance of negative pairs.
To measure the similarity of instance pairs, the cosine similarity can be used, that is
\begin{equation}\label{}
s(u, v)=\frac{u^{\top}v}{\|u\|\|v\|},
\end{equation}
where $u$ and $v$ denote two feature vectors.
To optimize the agreement of contrastive pairs constructed from two augmentations, say, augmentation $a$ and augmentation $b$, the contrastive loss for an augmented sample $v_{i}^{a}$ is defined as
\begin{equation}\label{}
\ell_{i}^{a}=-\log \frac{\exp (s(y_{i}^{a}, y_{i}^{b}) / \tau_{g})}{\sum_{j=1}^{N}[\exp (s(y_{i}^{a}, y_{j}^{a}) / \tau_{g})+\exp (s(y_{i}^{a}, y_{j}^{b}) / \tau_{g})]},
\end{equation}
with $i,j \in[1, N]$ and $a,b \in\{1, 2, 3\}$. The parameter $\tau_{g}$ is the temperature parameter. In order to identify all positive pairs from two augmentations (say, $a$ and $b$), the instance-level contrastive loss is calculated over every augmented examples, that is
\begin{equation}\label{}
\mathcal{L}_{instance(a, b)}=\frac{1}{2 N} \sum_{i=1}^{N}(\ell_{i}^{a}+\ell_{i}^{b})
\end{equation}
In the instance projector, we construct two augmented view pairs for each original input. One pair consists of a strongly augmented view and a weakly augmented view, while the other one consists of two weakly augmented views. Thus the contrastive loss for the instance projector is defined as
\begin{equation}\label{}
\mathcal{L}_{instance}=\mathcal{L}_{instance(1, 2)}+\mathcal{L}_{instance(2, 3)}
\end{equation}

\subsubsection{Cluster Projector with Weak-Strong Augmentations}

The cluster projector is a two-layer nonlinear MLP with a softmax layer, denoted as $h(\cdot)$. The dimension of the output layer of the cluster projector, denoted as $M$, is equal to the number of classes (or the desired number of clusters). The output representation (for each sample) in the cluster projector, computed by $\tilde{c}_{i}^{j} = h(z_{i}^{j})$, can be treated as the probabilities of this sample belonging to different classes. Thus, $\tilde{c}_{i}^{j}$ can serve as a soft label for the augmented sample.

For each of the three augmentation views, a feature matrix with $N$ rows and $M$ columns can be obtained for a mini-batch of $N$ samples, where $c_{m}^{j}$ denotes the $m$-th column of the feature matrix, with $m \in[1, M]$ and $j \in\{1, 2, 3\}$. That is, $c_{m}^{j}$ can be regarded as the distribution of the $N$ samples in the $m$-th cluster of the augmentation $j$. We treat the same cluster from two different augmentation views as a positive cluster pair, and the other cluster pairs as the negative cluster pairs.
Then, for a cluster $c_{m}^{j}$, the cluster-level contrastive loss can be defined as
\begin{equation}\label{}
\hat{\ell}_{m}^{a}=-\log \frac{\exp (s(c_{m}^{a}, c_{m}^{b}) / \tau_{h})}{\sum_{n=1}^{M}[\exp (s(c_{m}^{a}, c_{n}^{a}) / \tau_{h})+\exp (s(c_{m}^{a}, c_{n}^{b}) / \tau_{h})]}
\end{equation}
with $m,n \in[1, N]$ and $a,b \in\{1, 2, 3\}$. The parameter $\tau_{h}$ is the temperature parameter. After traversing all clusters, the cluster-level contrastive loss can further be represented as
\begin{align}\label{}
\mathcal{L}_{cluster(a, b)}=\frac{1}{2 M} \sum_{m=1}^{M}(\hat{\ell}_{m}^{a}+\hat{\ell}_{m}^{b})-H(Y),\\
H(Y)=-\sum_{m=1}^{M}[P(c_{m}^{a}) \log P(c_{m}^{a})+P(c_{m}^{b}) \log P(c_{m}^{b})],
\end{align}
where $H(Y)$ is the entropy of the cluster assignment probabilities with $P(c_{m}^{k})=\sum_{n=1}^{N} Y_{n m}^{k} /\|Y\|_{1}$, for $k \in\{a, b\}$ within a mini-batch under each data augmentation. This term is incorporated to avoid the trivial solution that most samples are assigned to the same cluster.

For every original image, we utilize three augmentation view pairs in the cluster projector, i.e., every two augmentation views form a view pair. Then the contrastive loss  in the cluster projector is defined as
\begin{equation}\label{}
\mathcal{L}_{cluster}=\mathcal{L}_{cluster(1, 2)}+\mathcal{L}_{cluster(1, 3)}+\mathcal{L}_{cluster(2, 3)}
\end{equation}

\subsubsection{Overall Objective}

The optimization of the backbone network, the instance projector, and the cluster projector is jointly performed in an end-to-end manner. The overall objective function is composed of the instance-level contrastive loss and the cluster-level contrastive loss, that is
\begin{equation}\label{}
\mathcal{L}=\mathcal{L}_{cluster}+\mathcal{L}_{instance} .
\end{equation}
Thereby, the unsupervised network training of our SACC approach can be conducted with both instance-level and cluster-level contrastive learning upon three weak/strong augmentation views.

\subsection{Implementation Details}
\label{sec:implement_details}
In the proposed framework, all original images of different sizes are resized to the size of 224$\times$224. We use the ResNet-34 as the backbone network which is designed for input images with 224$\times$224 pixels. As for the instance projector, its output dimensionality set to 128 in order to hold sufficient information after the transformation. As for the cluster projector,  the output dimensionality is set to the desired number of clusters, where the output feature vector can be treated as the soft label. The temperature parameters of the instance projector and the cluster projector are fixed to 0.5 and 1, respectively. In the training process, we adopt Adam optimizer with a learning rate of 0.0003 to simultaneously optimize the backbone network and the two projectors. The batch size for training is set to 200, and the number of training epochs is set to 1000.

\begin{table}[!t]
	\renewcommand\arraystretch{1.1}
	\begin{center}
		\begin{minipage}{220pt}
			\caption{Description of the benchmark datasets.}
			\label{tab1}
			\begin{tabular}{p{2.88cm}<{\centering}p{1.875cm}<{\centering}p{1.875cm}<{\centering}}
				\toprule
				Dataset       & \#Images & \#Classes \\
				\midrule
				CIFAR-10      & 60,000       & 10       \\
				CIFAR-100     & 60,000       & 20       \\
				STL-10        & 13,000       & 10       \\
				ImageNet-10   & 13,000       & 10       \\
				ImageNet-Dogs & 19,500       & 15       \\
				\bottomrule
			\end{tabular}
		\end{minipage}
	\end{center}
\end{table}

\begin{figure*}[!t]\vskip 0.1 in
\centering  
\subfigure[CIFAR-10]{
\label{Fig.sub.1}
\includegraphics[width=0.18\textwidth]{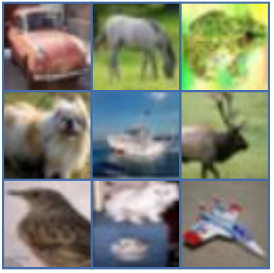}}
\subfigure[CIFAR-100]{
\label{Fig.sub.2}
\includegraphics[width=0.18\textwidth]{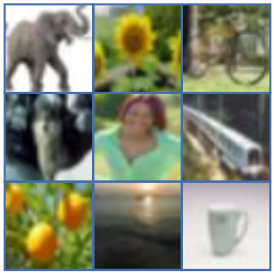}}
\subfigure[STL-10]{
\label{Fig.sub.3}
\includegraphics[width=0.18\textwidth]{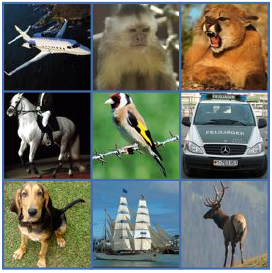}}
\subfigure[ImageNet-10]{
\label{Fig.sub.4}
\includegraphics[width=0.18\textwidth]{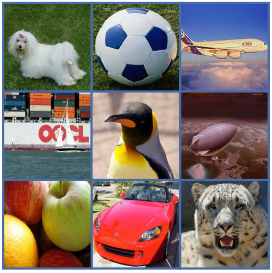}}
\subfigure[ImageNet-Dogs]{
\label{Fig.sub.5}
\includegraphics[width=0.18\textwidth]{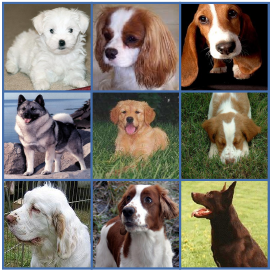}}
\caption{Visualization of some image samples from the five benchmark datasets.}\vskip 0.1 in
\label{fig3}
\end{figure*}

\begin{table*}[!t]
	\renewcommand\arraystretch{1}
	\centering
	\caption{The clustering performance (w.r.t. NMI(\%)) by different clustering algorithms on the five image datasets. The best score in each column is in \textbf{bold}.}
	\label{table:comp_nmi}
	\begin{tabular}{|p{3cm}<{\centering}||p{2cm}<{\centering}|p{2.1cm}<{\centering}|p{2cm}<{\centering}|p{2.1cm}<{\centering}|p{2.4cm}<{\centering}|}
		\hline
		Dataset            & CIFAR-10      & CIFAR-100     & STL-10        & ImageNet-10   & ImageNet-Dogs \\ \hline \hline
		K-means \cite{macqueen1967some}           & 8.7           & 8.4           & 12.5          & 11.9          & 5.5           \\ \hline
		SC \cite{zelnik2004self}                  & 10.3          & 9.0           & 9.8           & 15.1          & 3.8           \\ \hline
		AC \cite{gowda1978agglomerative}          & 10.5          & 9.8           & 23.9          & 13.8          & 3.7           \\ \hline
		NMF \cite{cai2009locality}                & 8.1           & 7.9           & 9.6           & 13.2          & 4.4           \\ \hline
		AE \cite{bengio2006greedy}                & 23.9          & 10.0          & 25.0          & 21.0          & 10.4          \\ \hline
		DAE \cite{vincent2010stacked}             & 25.1          & 11.1          & 22.4          & 20.6          & 10.4          \\ \hline
		DCGAN \cite{radford2015unsupervised}      & 26.5          & 12.0          & 21.0          & 22.5          & 12.1          \\ \hline
		DeCNN \cite{zeiler2010deconvolutional}    & 24.0          & 9.2           & 22.7          & 18.6          & 9.8           \\ \hline
		VAE \cite{kingma2013auto}                 & 24.5          & 10.8          & 20.0          & 19.3          & 10.7          \\ \hline
		JULE \cite{yang2016joint}                 & 19.2          & 10.3          & 18.2          & 17.5          & 5.4           \\ \hline
		DEC \cite{xie2016unsupervised}            & 25.7          & 13.6          & 27.6          & 28.2          & 12.2          \\ \hline
		DAC \cite{chang2017deep}                  & 39.6          & 18.5          & 36.6          & 39.4          & 21.9          \\ \hline
		DDC \cite{chang2019deep}                  & 42.4          & -             & 37.1          & 43.3          & -             \\ \hline
		DCCM \cite{wu2019deep}                    & 49.6          & 28.5          & 37.6          & 60.8          & 32.1          \\ \hline
		IIC \cite{ji2019invariant}                & 51.1          & 22.5          & 49.6          & -             & -             \\ \hline
		GATCluster \cite{niu2020gatcluster}       & 49.6          & 28.5          & 44.6          & 59.4          & 28.1          \\ \hline
		PICA \cite{huang2020deep}                 & 59.1          & 31.0          & 61.1          & 80.2          & 35.2          \\ \hline
		DRC \cite{zhong2020deep}                  & 62.1          & 35.6          & 64.4          & 83.0          & 38.4          \\ \hline
		CC \cite{li2021contrastive}               & 68.1          & 42.4          & 67.4          & 86.2          & 40.1          \\ \hline
		\textbf{SACC(our)} & \textbf{76.5} & \textbf{44.8} & \textbf{69.1} & \textbf{87.7} & \textbf{45.5} \\
		\hline
	\end{tabular}\vskip 0.1 in
\end{table*}

\section{Experiments}\label{sec4}

In this section, we conduct extensive experiments on five image datasets to evaluate the clustering performance of our SACC algorithm against eighteen traditional and deep clustering algorithms.

\subsection{Datasets and Evaluation Metrics}

In our experiments, five challenging image datasets are used, namely, CIFAR-10 \cite{krizhevsky2009learning}, CIFAR-100 \cite{krizhevsky2009learning}, STL-10 \cite{coates2011analysis}, ImageNet-10 \cite{chang2017deep}, and ImageNet-Dogs \cite{chang2017deep}. Similar to the previous deep clustering works \cite{wu2019deep,huang2020deep,li2021contrastive}, we jointly utilize the training and testing samples of each dataset, due to the unsupervised nature of the clustering task. Specifically, the five benchmark datasets are introduced below.

\begin{itemize}
  \item \textbf{CIFAR-10} is a natural image dataset which consists of 60,000 images from 10 object classes.
  \item \textbf{CIFAR-100} has the same size and the same number of samples as the CIFAR-10 dataset, but it contains 20 super-classes, which can be further divided into 100 classes. Following the previous works, we use the 20 super-classes as the ground-truth when evaluating the clustering performance.
  \item \textbf{STL-10} is an ImageNet-sourced dataset which collects 13,000 color images with the size of $96\times 96$ from 10 classes.
  \item \textbf{ImageNet-10} is a subset of ImageNet with 10 classes, each of which consists of 1,300 samples with varying image sizes.
  \item \textbf{ImageNet-Dogs} is constructed in a similar way to ImageNet-10, but it selects a total of 19,500 dog images of 15 breeds from the ImageNet dataset.
\end{itemize}
For clarity, we visualize some samples from the five image datasets in Fig.~\ref{fig3}, and provide the statistics of them in Table~\ref{tab1}.

To evaluate the clustering results of different clustering algorithms, we adopt three widely-used evaluation metrics \cite{wu2019deep,huang20_tkde,li2021contrastive}, namely,  clustering accuracy (ACC), normalized mutual information (NMI) and adjusted rand index (ARI). Note that greater values of the three metrics indicate better clustering results.

\begin{table*}[!t]
	\renewcommand\arraystretch{1}
	\centering
	\caption{The clustering performance (w.r.t. ACC(\%)) by different clustering algorithms on the five image datasets. The best score in each column is in \textbf{bold}.}
	\label{table:comp_acc}
	\begin{tabular}{|p{3cm}<{\centering}||p{2cm}<{\centering}|p{2.1cm}<{\centering}|p{2cm}<{\centering}|p{2.1cm}<{\centering}|p{2.4cm}<{\centering}|}
		\hline
		Dataset            & CIFAR-10      & CIFAR-100     & STL-10        & ImageNet-10   & ImageNet-Dogs \\ \hline
		\hline
		K-means \cite{macqueen1967some}           & 22.9          & 13.0          & 19.2          & 24.1          & 10.5          \\ \hline
		SC \cite{zelnik2004self}                  & 24.7          & 13.6          & 15.9          & 27.4          & 11.1          \\ \hline
		AC \cite{gowda1978agglomerative}          & 22.8          & 13.8          & 33.2          & 24.2          & 13.9          \\ \hline
		NMF \cite{cai2009locality}                & 19.0          & 11.8          & 18.0          & 23.0          & 11.8          \\ \hline
		AE \cite{bengio2006greedy}                & 31.4          & 16.5          & 30.3          & 31.7          & 18.5          \\ \hline
		DAE \cite{vincent2010stacked}             & 29.7          & 15.1          & 30.2          & 30.4          & 19.0          \\ \hline
		DCGAN \cite{radford2015unsupervised}      & 31.5          & 15.3          & 29.8          & 34.6          & 17.4          \\ \hline
		DeCNN \cite{zeiler2010deconvolutional}    & 28.2          & 13.3          & 29.9          & 31.3          & 17.5          \\ \hline
		VAE \cite{kingma2013auto}                 & 29.1          & 15.2          & 28.2          & 33.4          & 17.9          \\ \hline
		JULE \cite{yang2016joint}                 & 27.2          & 13.7          & 27.7          & 30.0          & 13.8          \\ \hline
		DEC \cite{xie2016unsupervised}            & 30.1          & 18.5          & 35.9          & 38.1          & 19.5          \\ \hline
		DAC \cite{chang2017deep}                  & 52.2          & 23.8          & 47.0          & 52.7          & 27.5          \\ \hline
		DDC \cite{chang2019deep}                  & 52.4          & -             & 48.9          & 57.7          & -             \\ \hline
		DCCM \cite{wu2019deep}                    & 62.3          & 32.7          & 48.2          & 71.0          & 38.3          \\ \hline
		IIC \cite{ji2019invariant}                & 61.7          & 25.7          & 59.6          & -             & -             \\ \hline
		GATCluster \cite{niu2020gatcluster}       & 62.3          & 32.7          & 58.3          & 73.9          & 32.2          \\ \hline
		PICA \cite{huang2020deep}                 & 69.6          & 33.7          & 71.3          & 87.0           & 35.2          \\ \hline
		DRC \cite{zhong2020deep}                  & 72.7          & 36.7          & 74.7          & 88.4          & 38.9          \\ \hline
		CC \cite{li2021contrastive}               & 76.6          & 42.6          & 74.7          & 89.5          & 34.2          \\ \hline
		\textbf{SACC(our)} & \textbf{85.1} & \textbf{44.3} & \textbf{75.9} & \textbf{90.5} & \textbf{43.7} \\
		\hline
	\end{tabular}
\end{table*}

\begin{table*}[!t]
	\renewcommand\arraystretch{1}
	\centering
	\caption{The clustering performance (w.r.t. ARI(\%)) by different clustering algorithms on the five image datasets. The best score in each column is in \textbf{bold}.}
	\label{table:comp_ari}
	\begin{tabular}{|p{3cm}<{\centering}||p{2cm}<{\centering}|p{2.1cm}<{\centering}|p{2cm}<{\centering}|p{2.1cm}<{\centering}|p{2.4cm}<{\centering}|}
		\hline
		Dataset            & CIFAR-10      & CIFAR-100     & STL-10        & ImageNet-10   & ImageNet-Dogs \\ \hline \hline
		K-means \cite{macqueen1967some}         & 4.9           & 2.8           & 6.1           & 5.7           & 2.0           \\ \hline
		SC \cite{zelnik2004self}                & 8.5           & 2.2           & 4.8           & 7.6           & 1.3           \\ \hline
		AC \cite{gowda1978agglomerative}        & 6.5           & 3.4           & 14.0          & 6.7           & 2.1           \\ \hline
		NMF \cite{cai2009locality}              & 3.4           & 2.6           & 4.6           & 6.5           & 1.6           \\ \hline
		AE \cite{bengio2006greedy}              & 16.9          & 4.8           & 16.1          & 15.2          & 7.3           \\ \hline
		DAE \cite{vincent2010stacked}           & 16.3          & 4.6           & 15.2          & 13.8          & 7.8           \\ \hline
		DCGAN \cite{radford2015unsupervised}    & 17.6          & 4.5           & 13.9          & 15.7          & 7.8           \\ \hline
		DeCNN \cite{zeiler2010deconvolutional}  & 17.4          & 3.8           & 16.2          & 14.2          & 7.3           \\ \hline
		VAE \cite{kingma2013auto}               & 16.7          & 4.0           & 14.6          & 16.8          & 7.9           \\ \hline
		JULE \cite{yang2016joint}               & 13.8          & 3.3           & 16.4          & 13.8          & 2.8           \\ \hline
		DEC \cite{xie2016unsupervised}          & 16.1          & 5.0           & 18.6          & 20.3          & 7.9           \\ \hline
		DAC \cite{chang2017deep}                & 30.6          & 8.8           & 25.7          & 30.2          & 11.1          \\ \hline
		DDC \cite{chang2019deep}                & 32.9          & -             & 26.7          & 34.5          & -             \\ \hline
		DCCM \cite{wu2019deep}                  & 40.8          & 17.3          & 26.2          & 55.5          & 18.2          \\ \hline
		IIC \cite{ji2019invariant}              & 41.1          & 11.7          & 39.7          & -             & -             \\ \hline
		GATCluster \cite{niu2020gatcluster}     & 40.8          & 17.3          & 36.3          & 55.2          & 16.3          \\ \hline
		PICA \cite{huang2020deep}               & 51.2          & 17.1          & 53.1          & 76.1           & 20.1          \\ \hline
		DRC \cite{zhong2020deep}                & 54.7          & 20.8          & 56.9          & 79.8          & 23.3          \\ \hline
		CC \cite{li2021contrastive}             & 60.6          & 26.7          & 60.6          & 82.5          & 22.5          \\ \hline
		\textbf{SACC(our)} & \textbf{72.4} & \textbf{28.2} & \textbf{62.6} & \textbf{84.3} & \textbf{28.5} \\
		\hline
	\end{tabular}\vskip 0.04 in
\end{table*}

\subsection{Baseline Methods}

In the experiments, the proposed SACC method is compared with both traditional clustering methods and deep clustering methods.  Specifically, eighteen baseline clustering methods are compared, including four traditional clustering methods, namely, $K$-means \cite{macqueen1967some}, agglomerative clustering (AC) \cite{gowda1978agglomerative}, spectral clustering (SC) \cite{zelnik2004self}, and nonnegative matrix factorization (NMF) \cite{cai2009locality}, and fourteen deep clustering methods, including autoencoders (AE) \cite{bengio2006greedy}, denoising autoencoder (DAE) \cite{vincent2010stacked}, deep convolutional generative adversarial networks (DCGAN) \cite{radford2015unsupervised}, deconvolutional networks (DeCNN) \cite{zeiler2010deconvolutional}, variational auto-encoder (VAE) \cite{kingma2013auto}, joint unsupervised learning (JULE) \cite{yang2016joint}, deep embedded clustering (DEC) \cite{xie2016unsupervised}, deep adaptive clustering (DAC) \cite{chang2017deep}, deep discriminative clustering (DDC) \cite{chang2019deep}, deep comprehensive correlation mining (DCCM) \cite{wu2019deep}, invariant information clustering (IIC) \cite{ji2019invariant}, Gaussian attention network for image clustering (GATCluster) \cite{niu2020gatcluster}, partition confidence maximization (PICA) \cite{huang2020deep}, deep robust clustering (DRC) \cite{zhong2020deep} and contrastive clustering (CC) \cite{li2021contrastive}. For the CC method, the NMI, ACC, and ARI scores are reproduced by running the authors' code \cite{li2021contrastive}, while the scores of the other baseline methods are taken from the corresponding papers.

\subsection{Results and Analysis}

\begin{figure*}[!t] \vskip 0.1 in
\centering
\subfigure[CIFAR-10]{
\includegraphics[width=0.1879\textwidth]{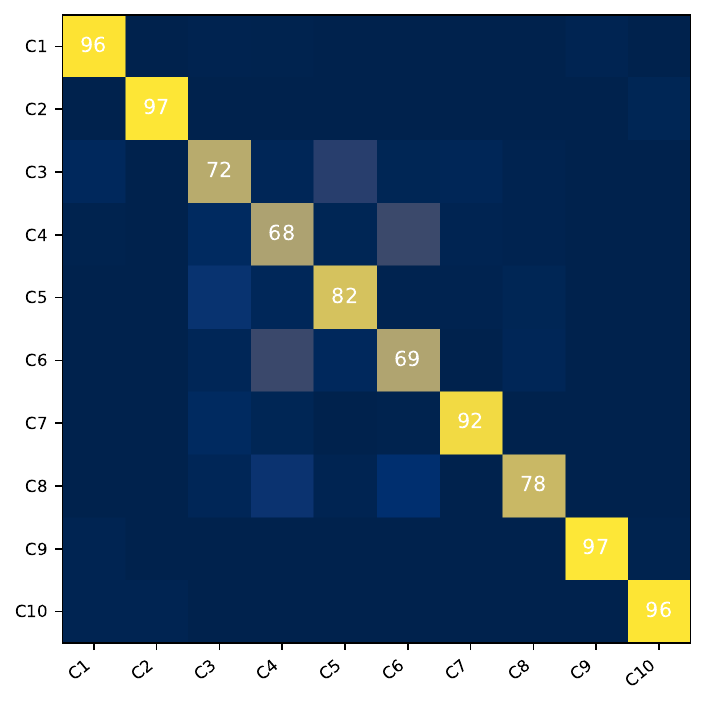}}
\subfigure[CIFAR-100]{
\includegraphics[width=0.1879\textwidth]{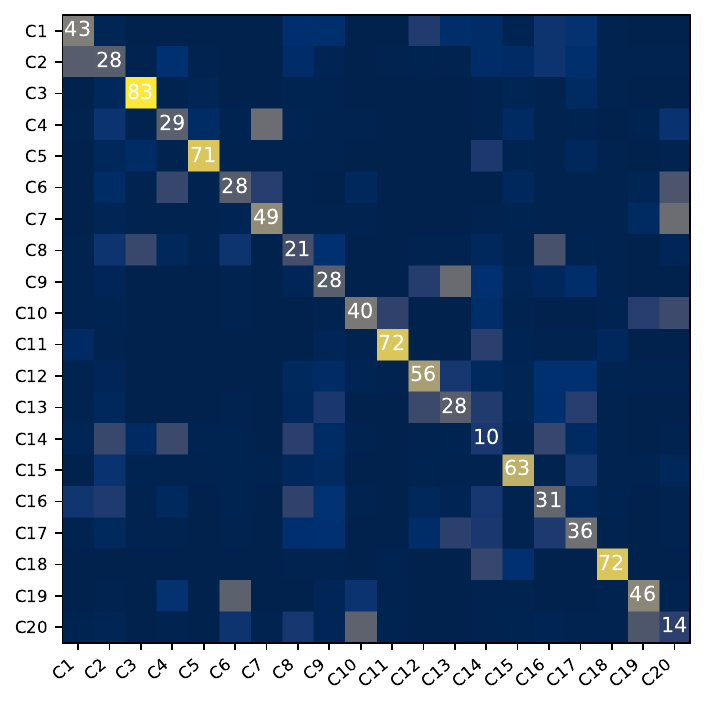}}
\subfigure[STL-10]{
\includegraphics[width=0.1879\textwidth]{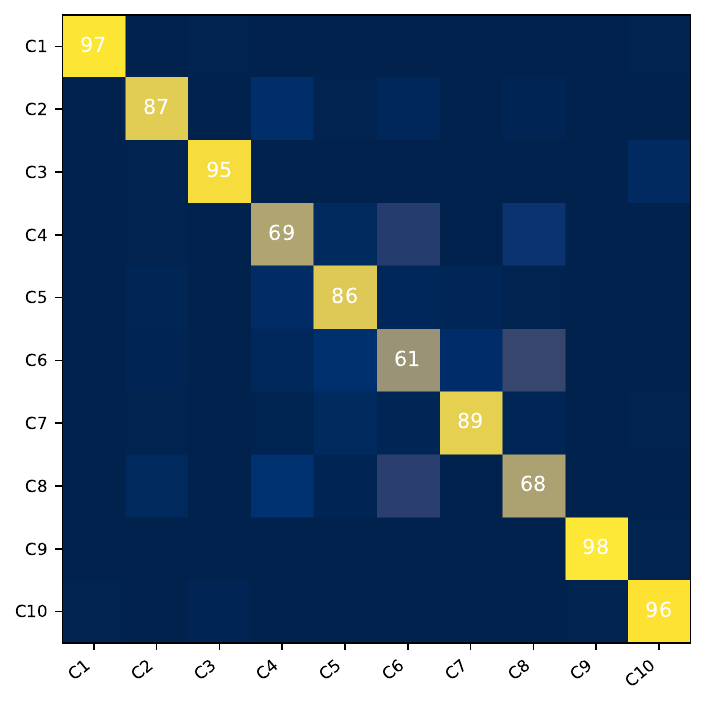}}
\subfigure[ImageNet-10]{
\includegraphics[width=0.1879\textwidth]{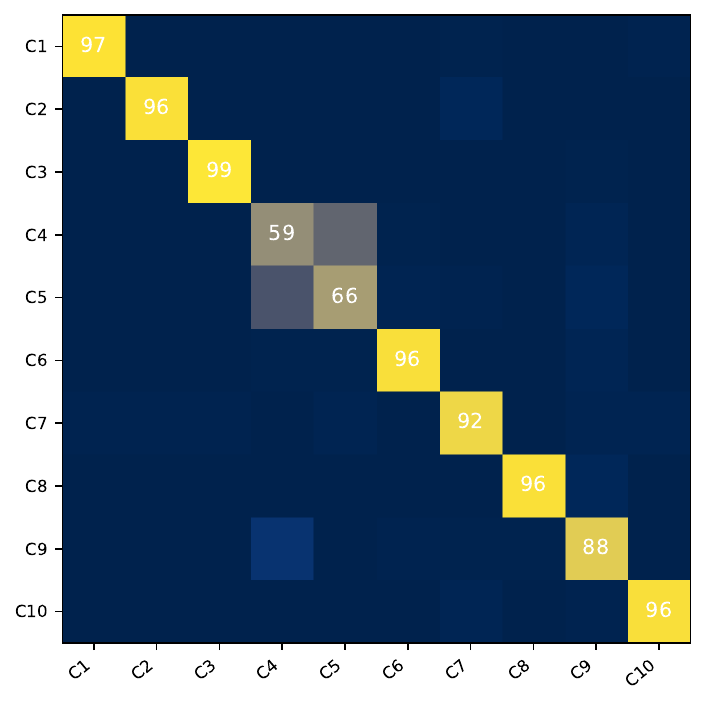}}
\subfigure[ImageNet-Dogs]{
\includegraphics[width=0.1879\textwidth]{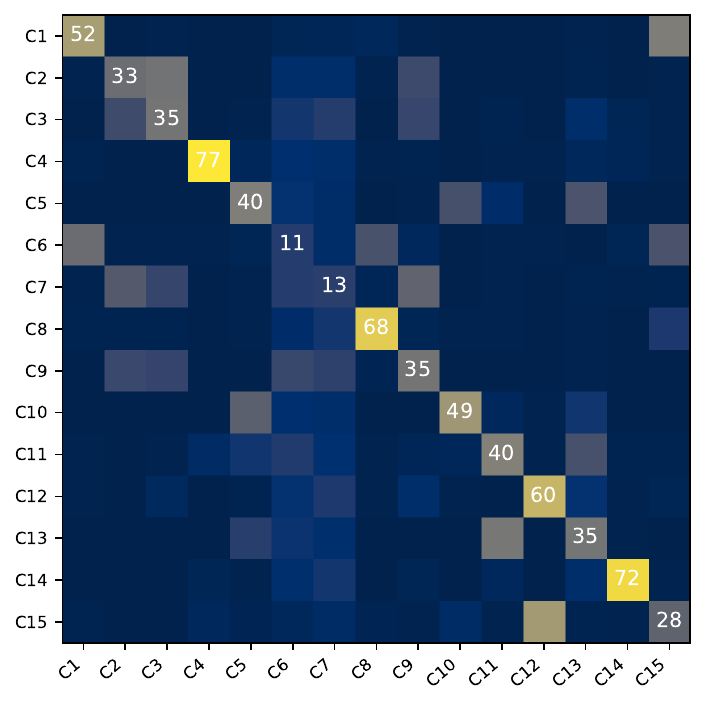}}
\caption{The confusion matrices on the five image datasets, where the rows are the ground-truth labels and the columns are the predicted labels by SACC.}
\label{fig5}\vskip 0.05 in
\end{figure*}

\begin{figure*}[!t]
\centering  
\subfigure[CIFAR-10]{
\includegraphics[width=0.18\textwidth]{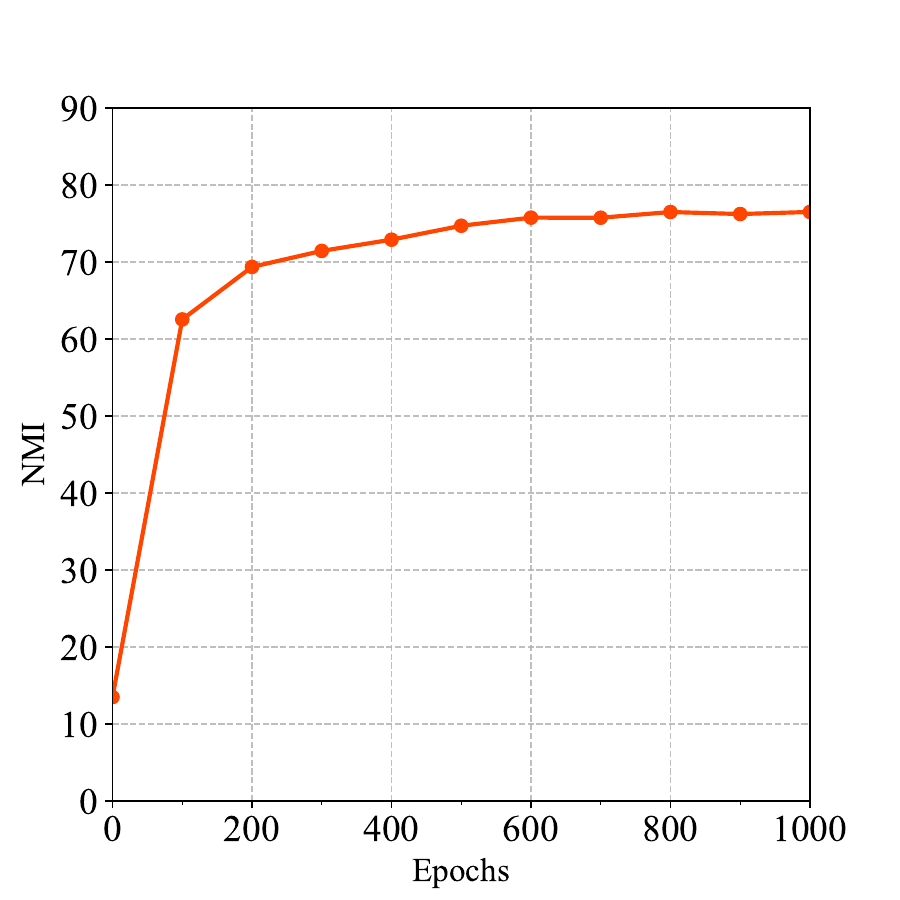}}
\subfigure[CIFAR-100]{
\includegraphics[width=0.18\textwidth]{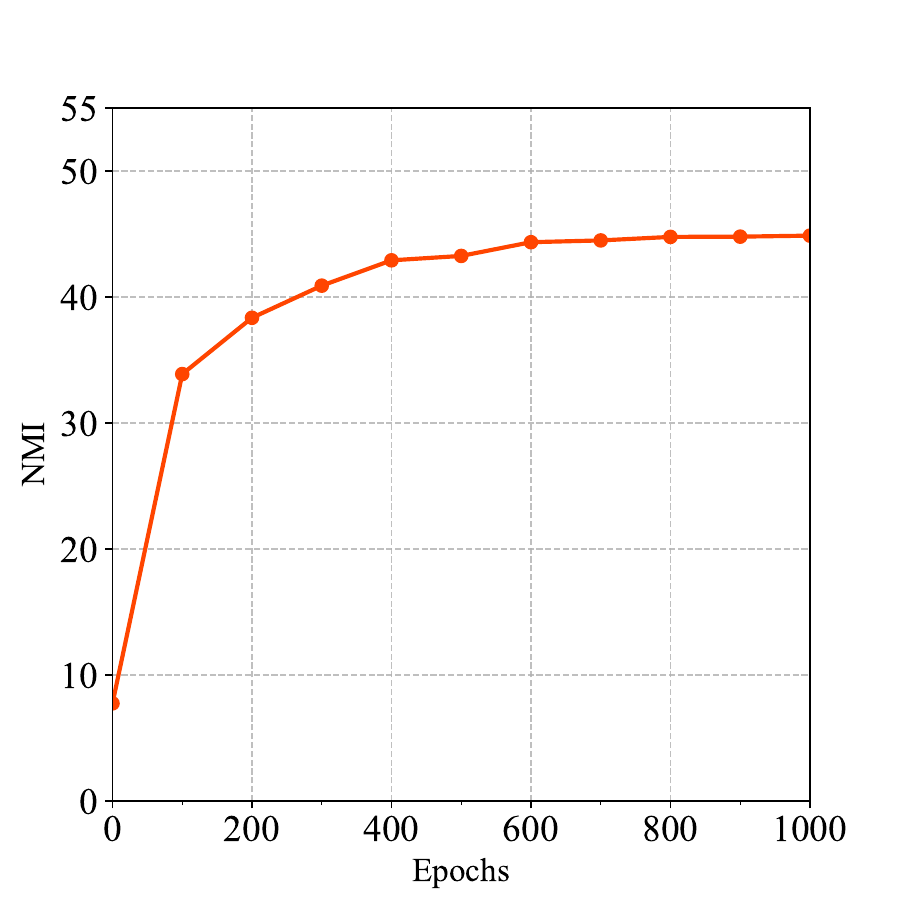}}
\subfigure[STL-10]{
\includegraphics[width=0.18\textwidth]{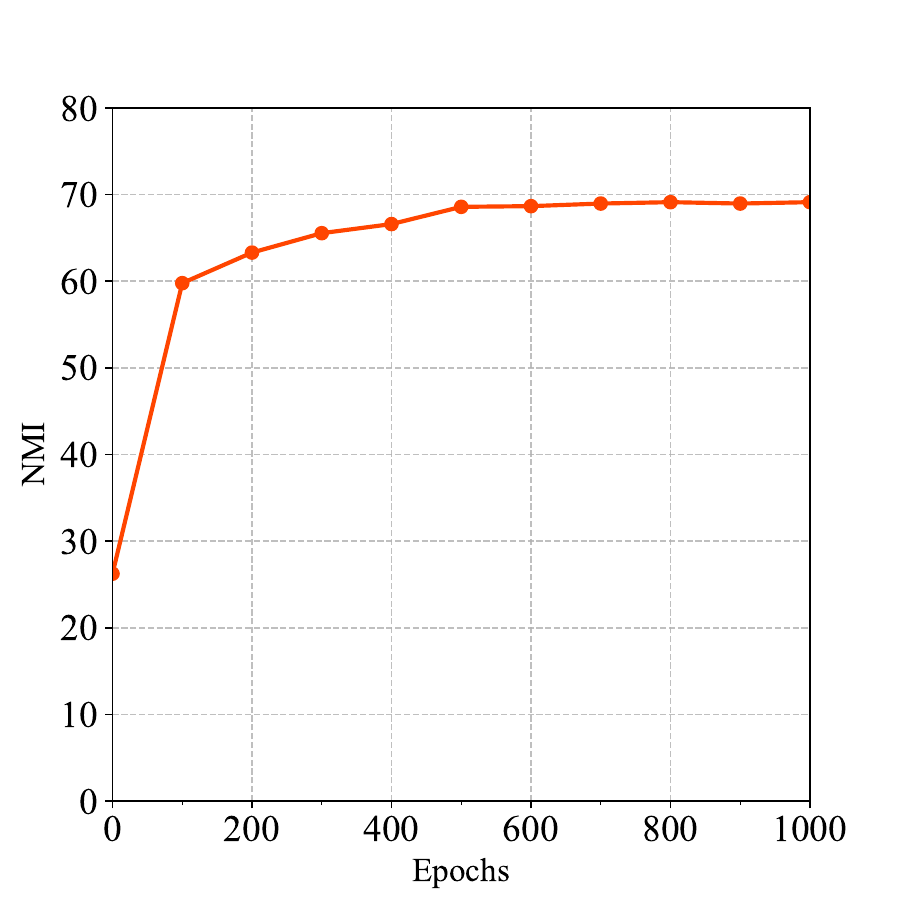}}
\subfigure[ImageNet-10]{
\includegraphics[width=0.18\textwidth]{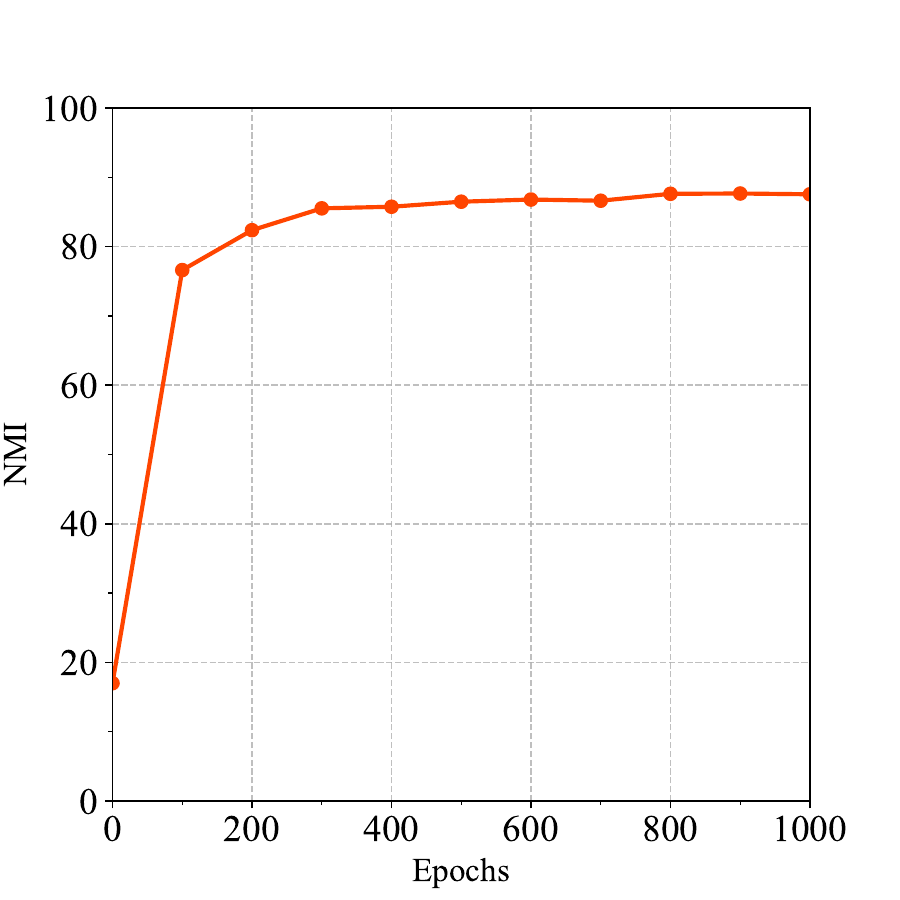}}
\subfigure[ImageNet-dogs]{
\includegraphics[width=0.18\textwidth]{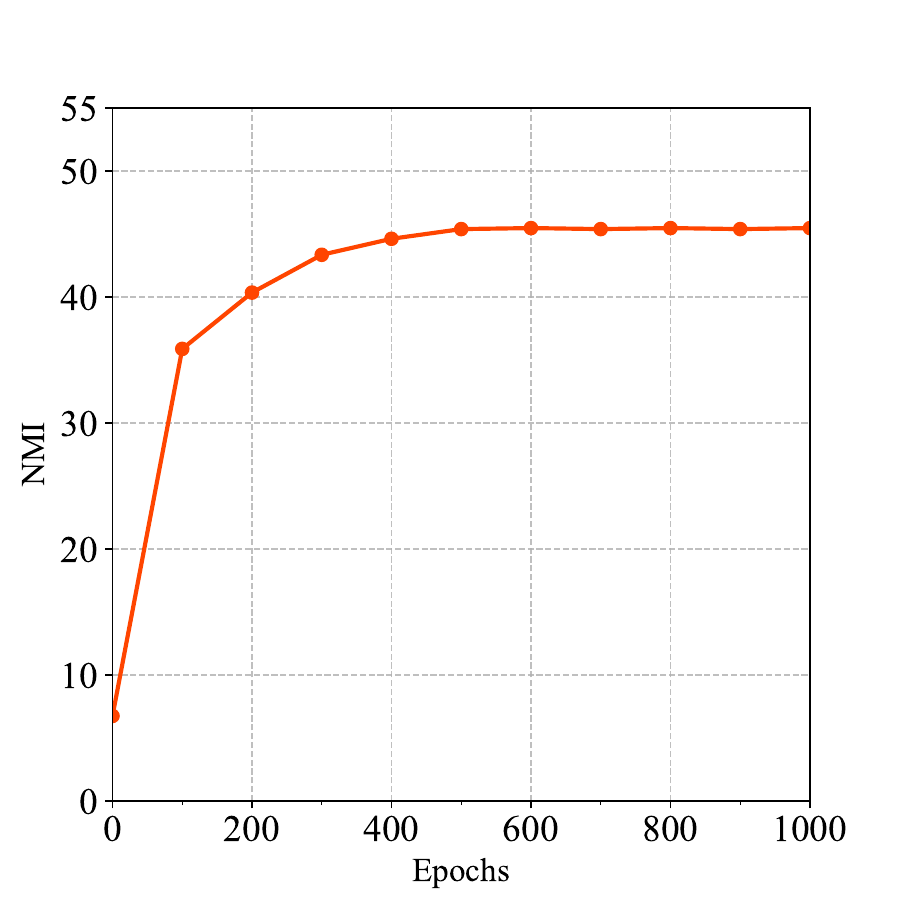}}
\caption{The NMI(\%) performance of SACC as the number of epochs increases.}
\label{fig6}
\end{figure*}

\begin{figure*}[!t]
\centering  
\subfigure[CIFAR-10]{
\includegraphics[width=0.18\textwidth]{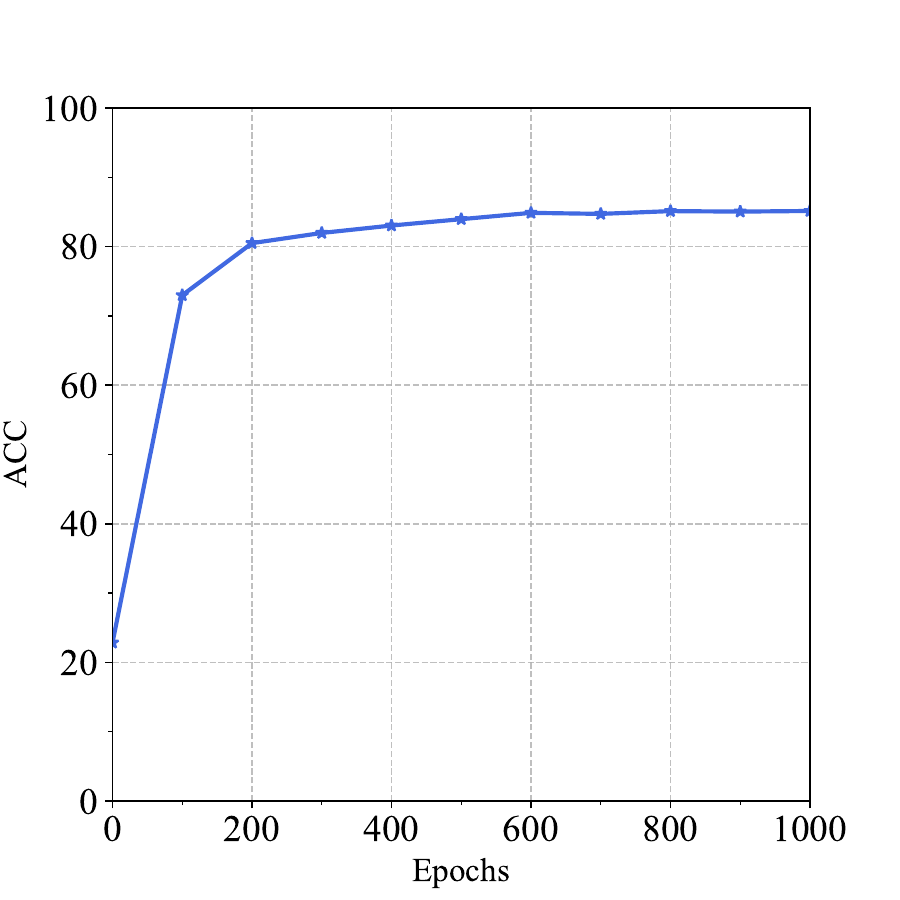}}
\subfigure[CIFAR-100]{
\includegraphics[width=0.18\textwidth]{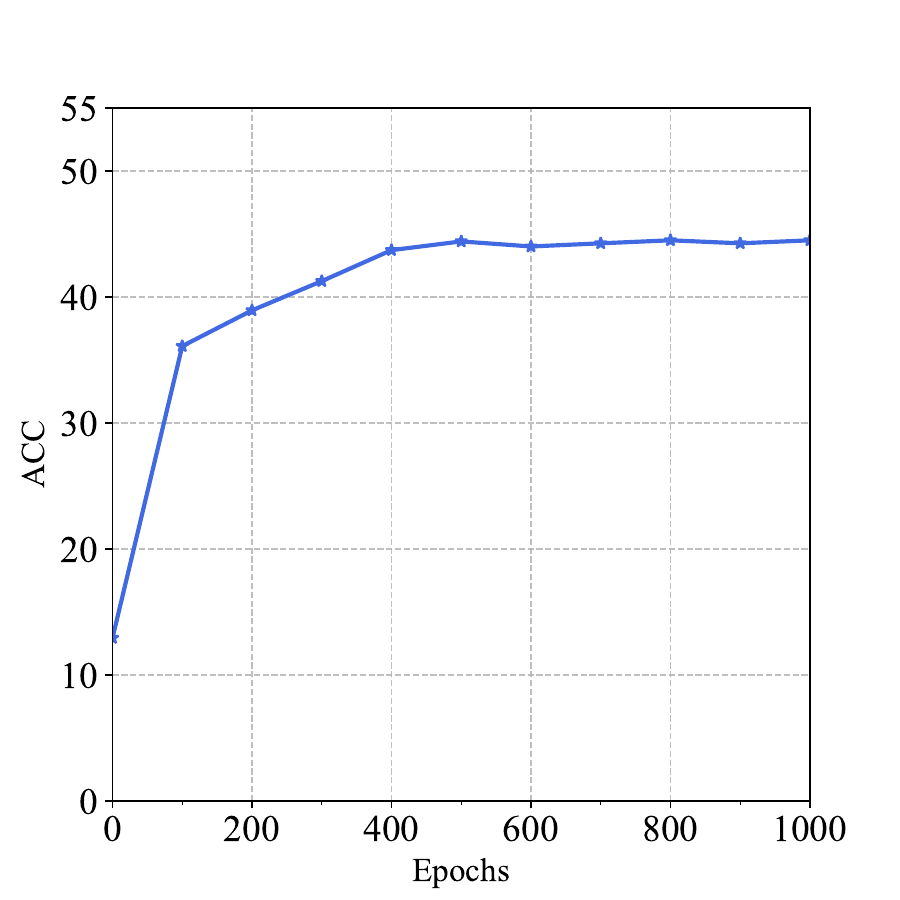}}
\subfigure[STL-10]{
\includegraphics[width=0.18\textwidth]{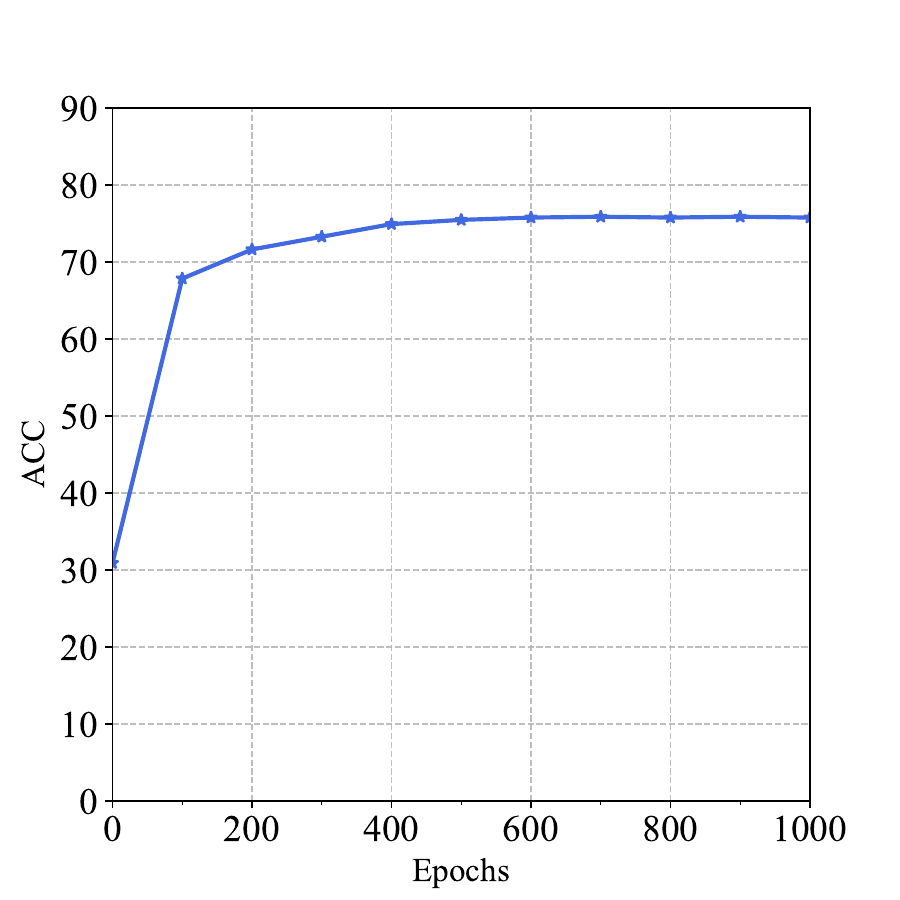}}
\subfigure[ImageNet-10]{
\includegraphics[width=0.18\textwidth]{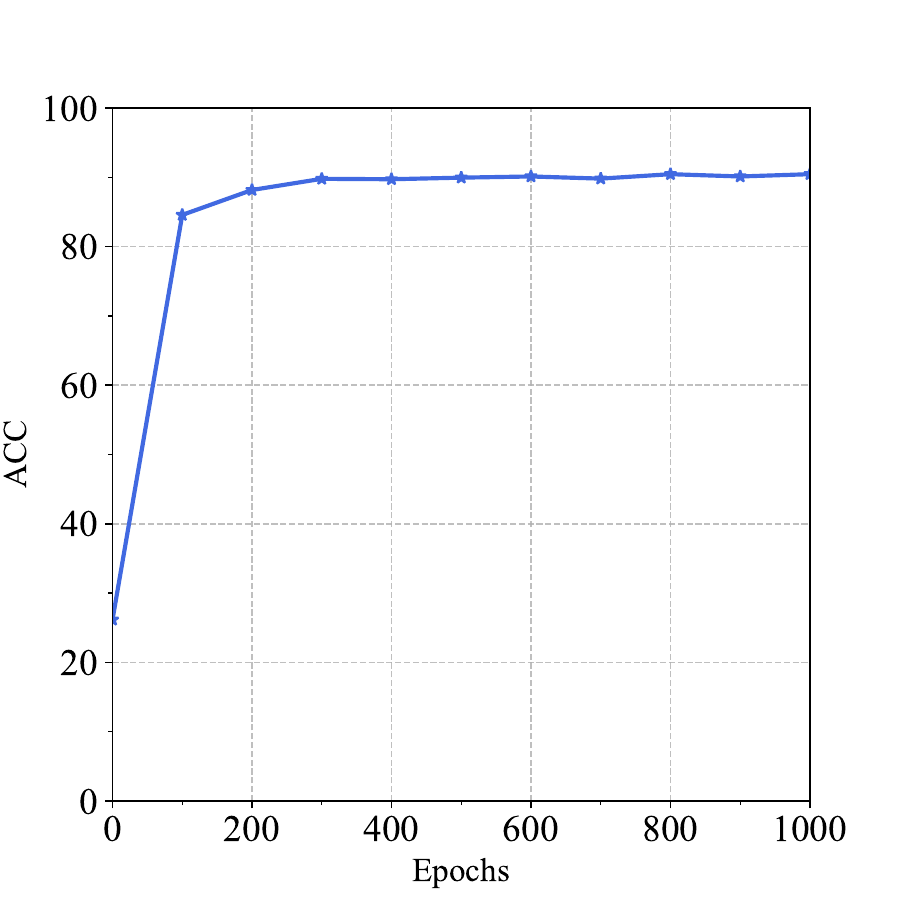}}
\subfigure[ImageNet-dogs]{
\includegraphics[width=0.18\textwidth]{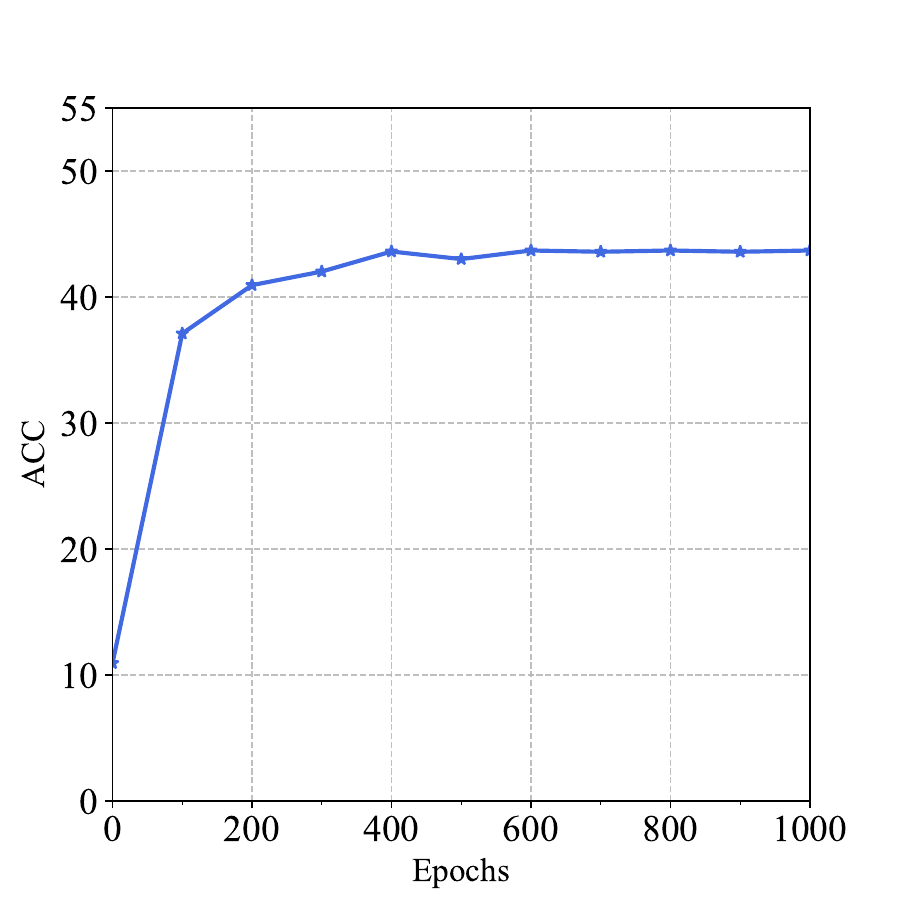}}
\caption{The ACC(\%) performance of SACC as the number of epochs increases.}
\label{fig7}
\end{figure*}

\begin{figure*}[!t]
\centering  
\subfigure[CIFAR-10]{
\includegraphics[width=0.18\textwidth]{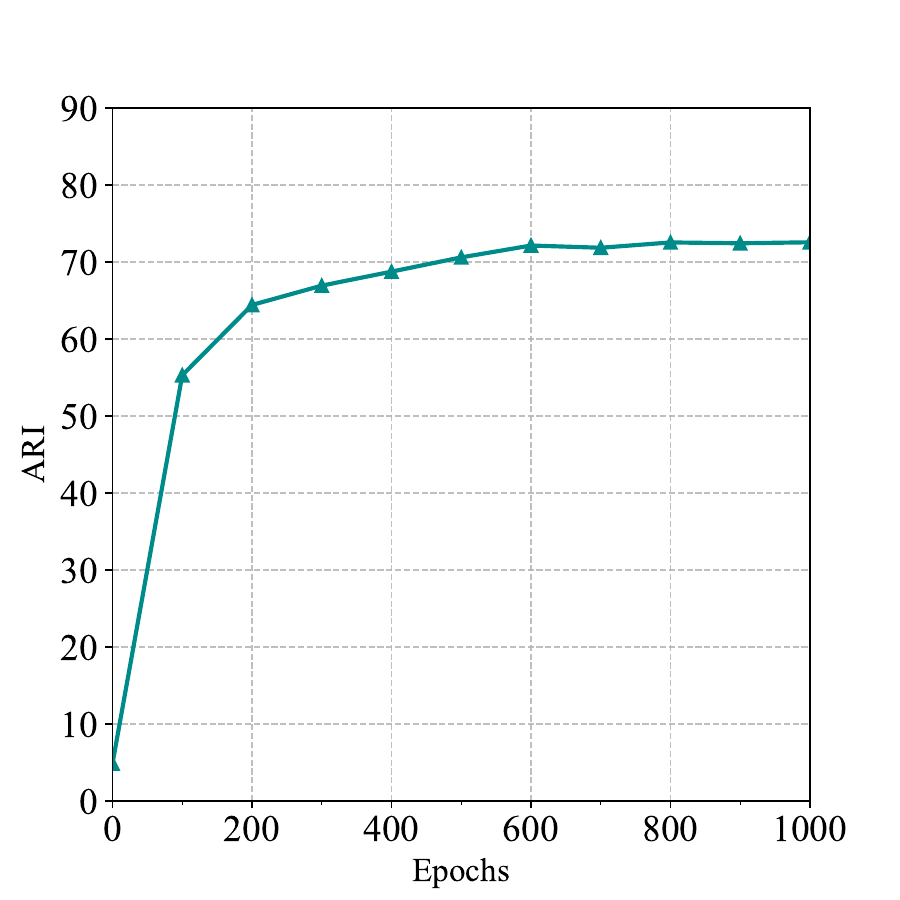}}
\subfigure[CIFAR-100]{
\includegraphics[width=0.18\textwidth]{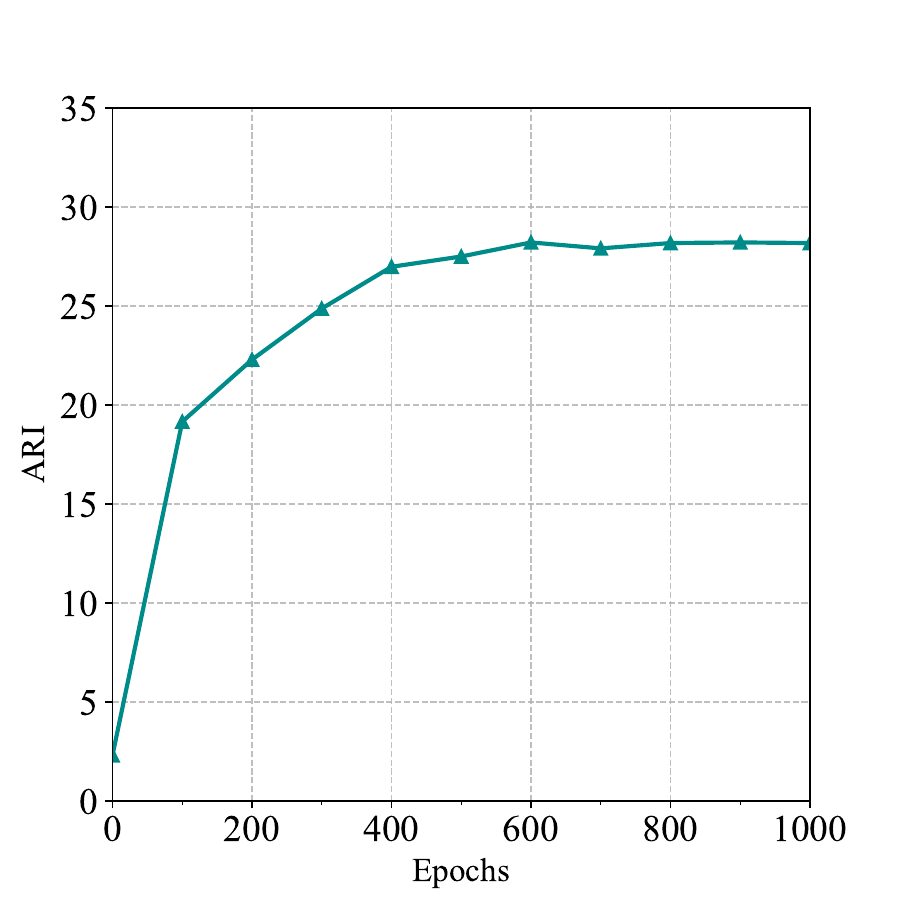}}
\subfigure[STL-10]{
\includegraphics[width=0.18\textwidth]{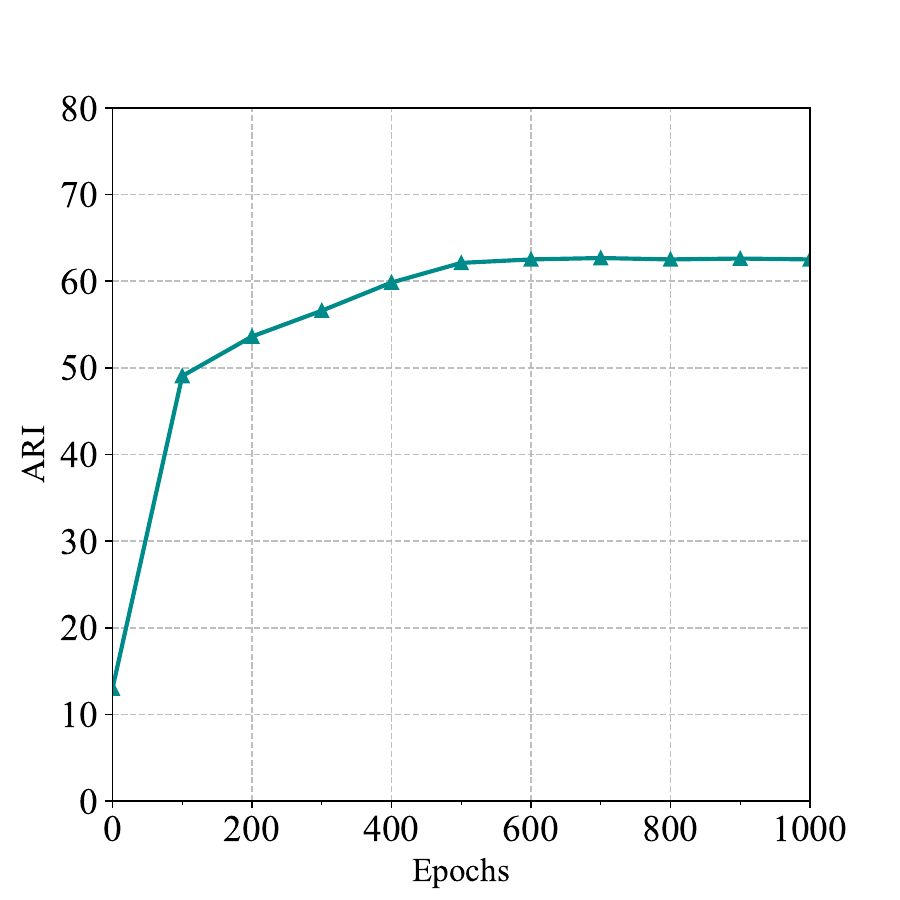}}
\subfigure[ImageNet-10]{
\includegraphics[width=0.18\textwidth]{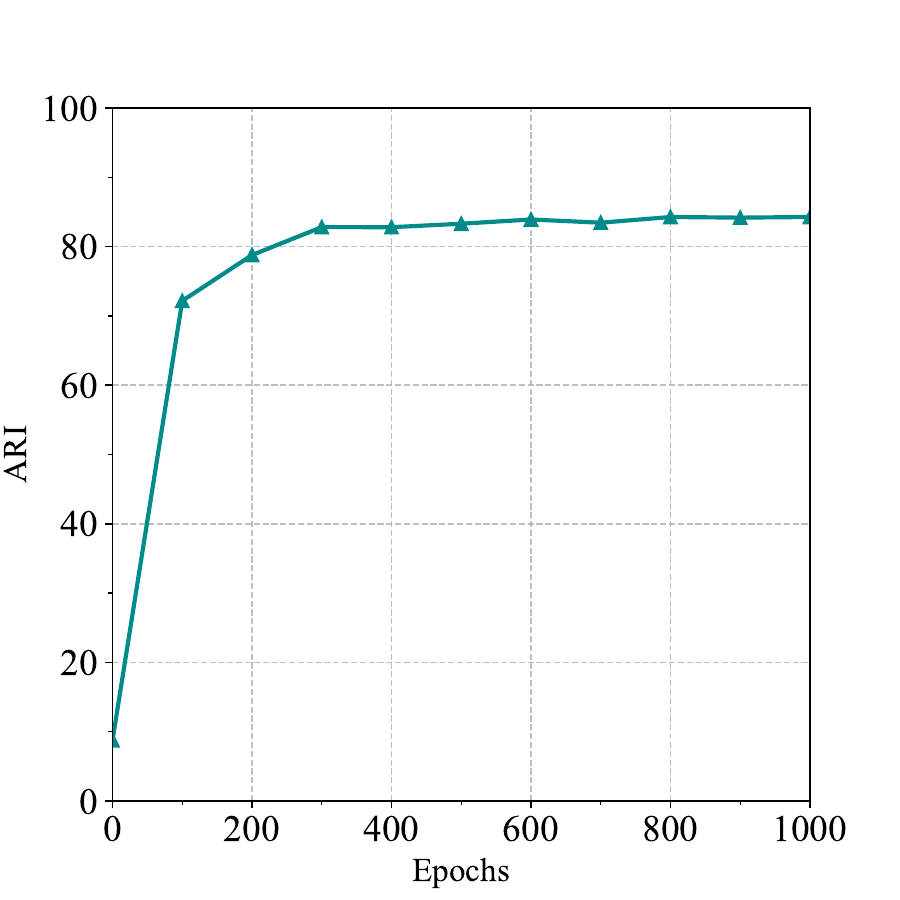}}
\subfigure[ImageNet-dogs]{
\includegraphics[width=0.18\textwidth]{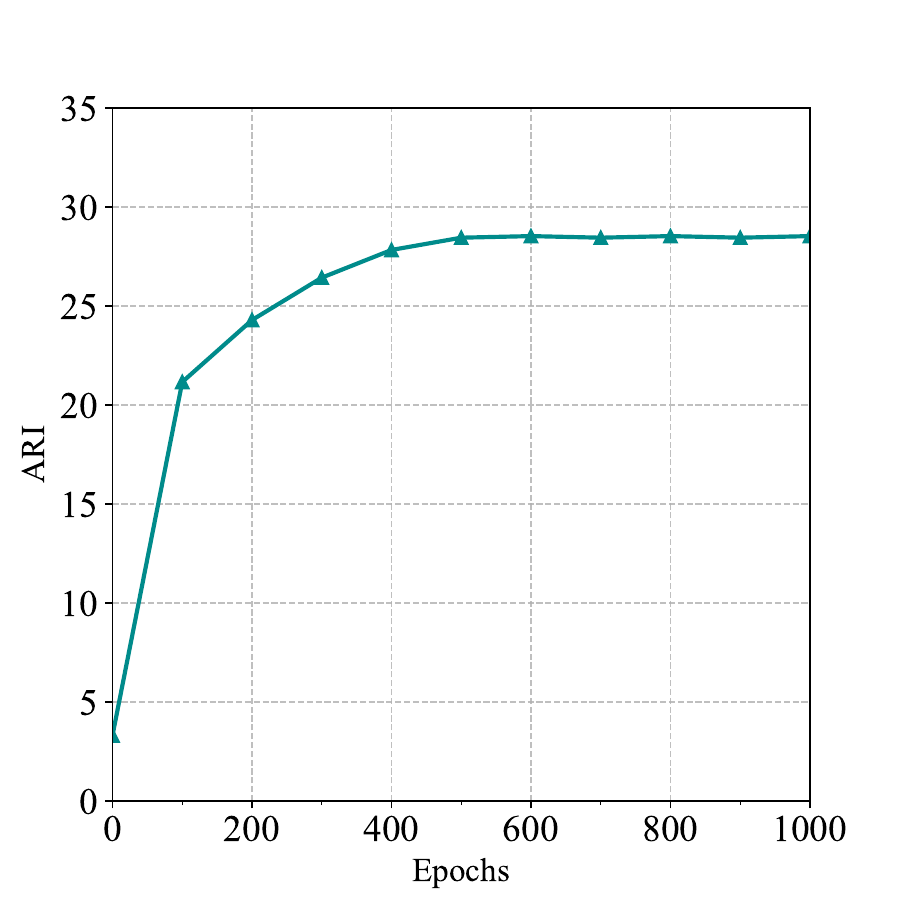}}
\caption{The ARI(\%) performance of SACC as the number of epochs increases.}\vskip 0.1 in
\label{fig8}
\end{figure*}

\begin{figure*}[!t]\vskip 0.1 in
\centering  
\subfigure[0 epoch]{
\includegraphics[width=0.3\textwidth]{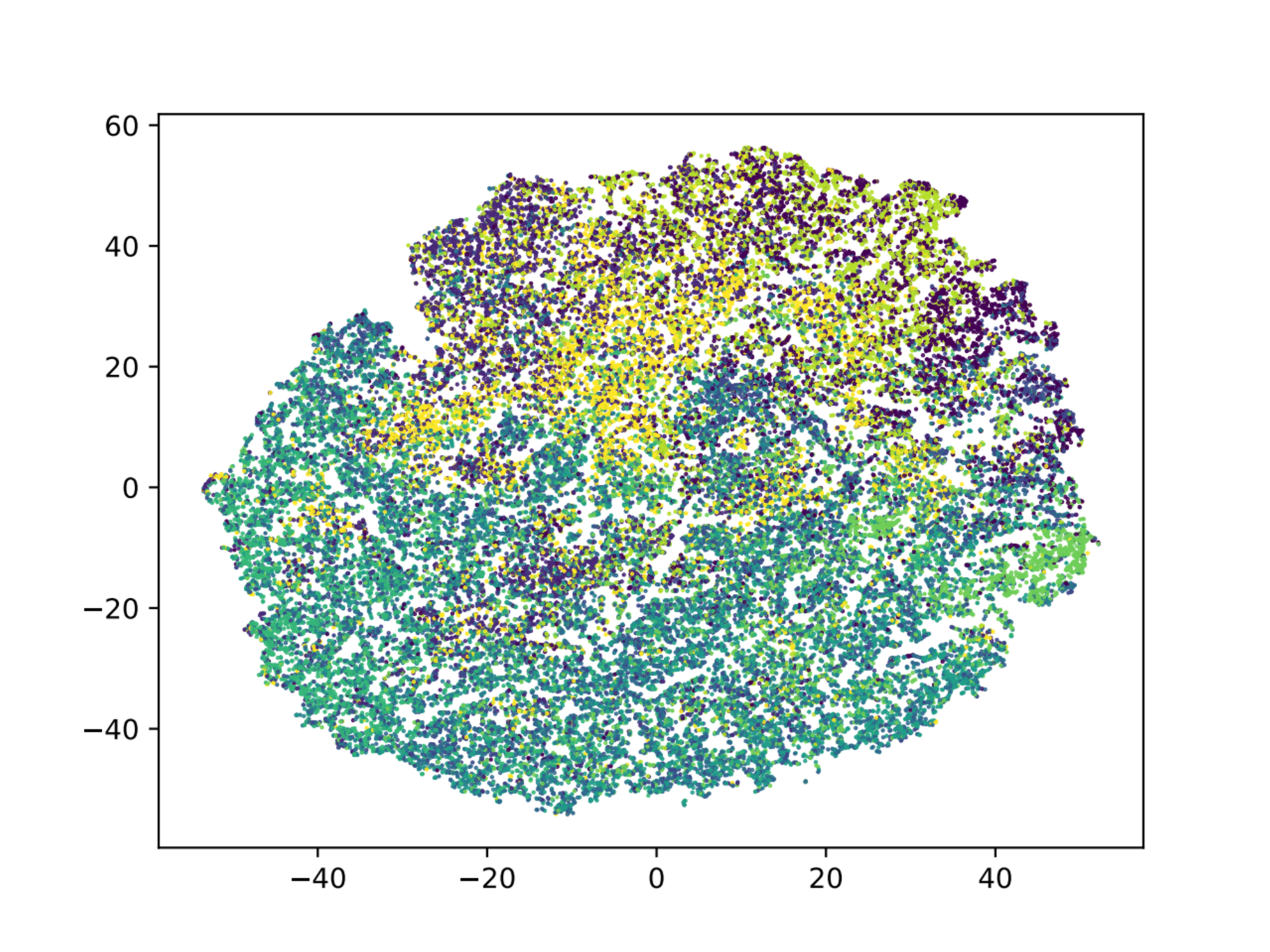}}
\subfigure[100 epoch]{
\includegraphics[width=0.3\textwidth]{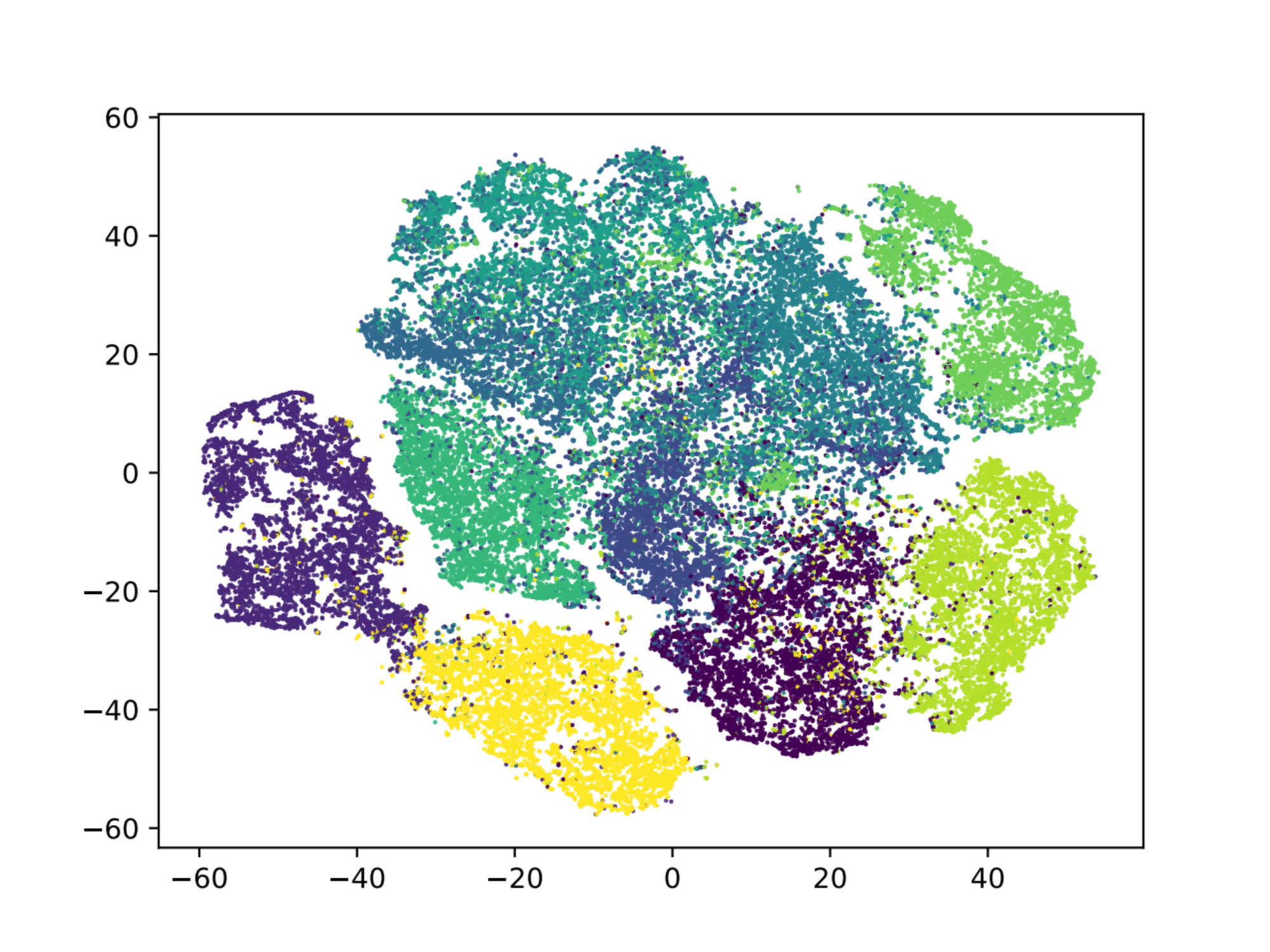}}
\subfigure[200 epoch]{
\includegraphics[width=0.3\textwidth]{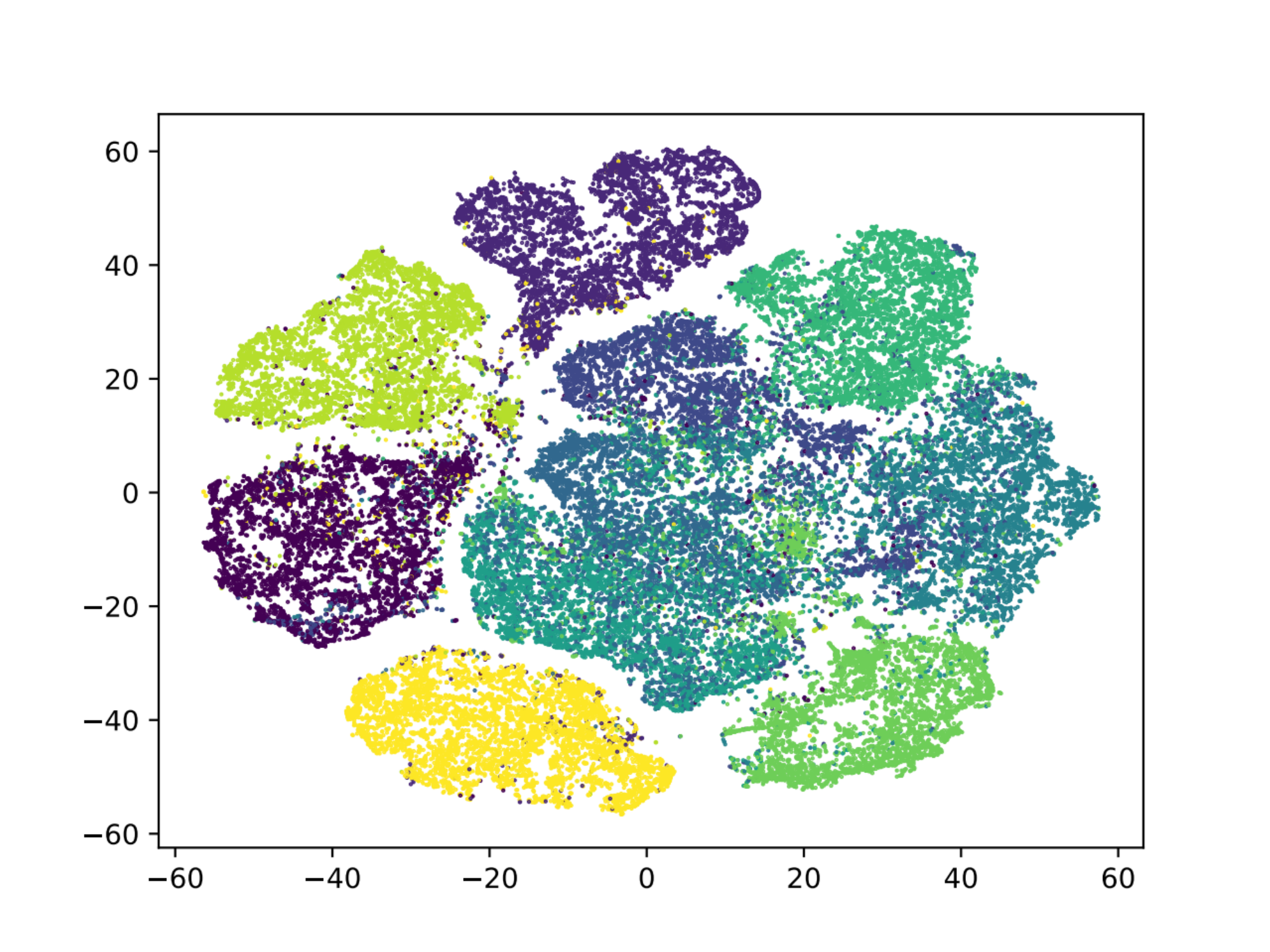}}
\subfigure[300 epoch]{
\includegraphics[width=0.3\textwidth]{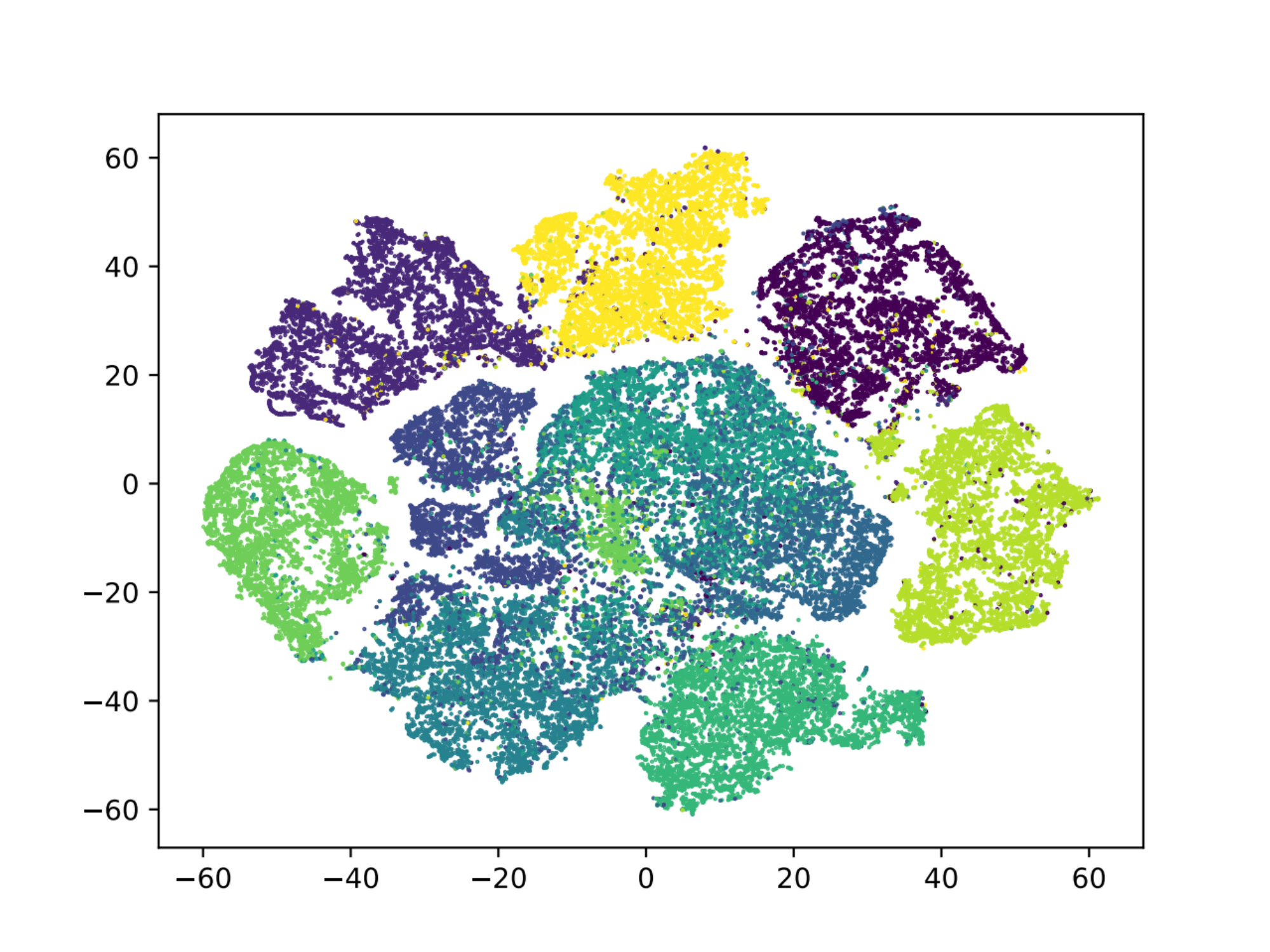}}
\subfigure[500 epoch]{
\includegraphics[width=0.3\textwidth]{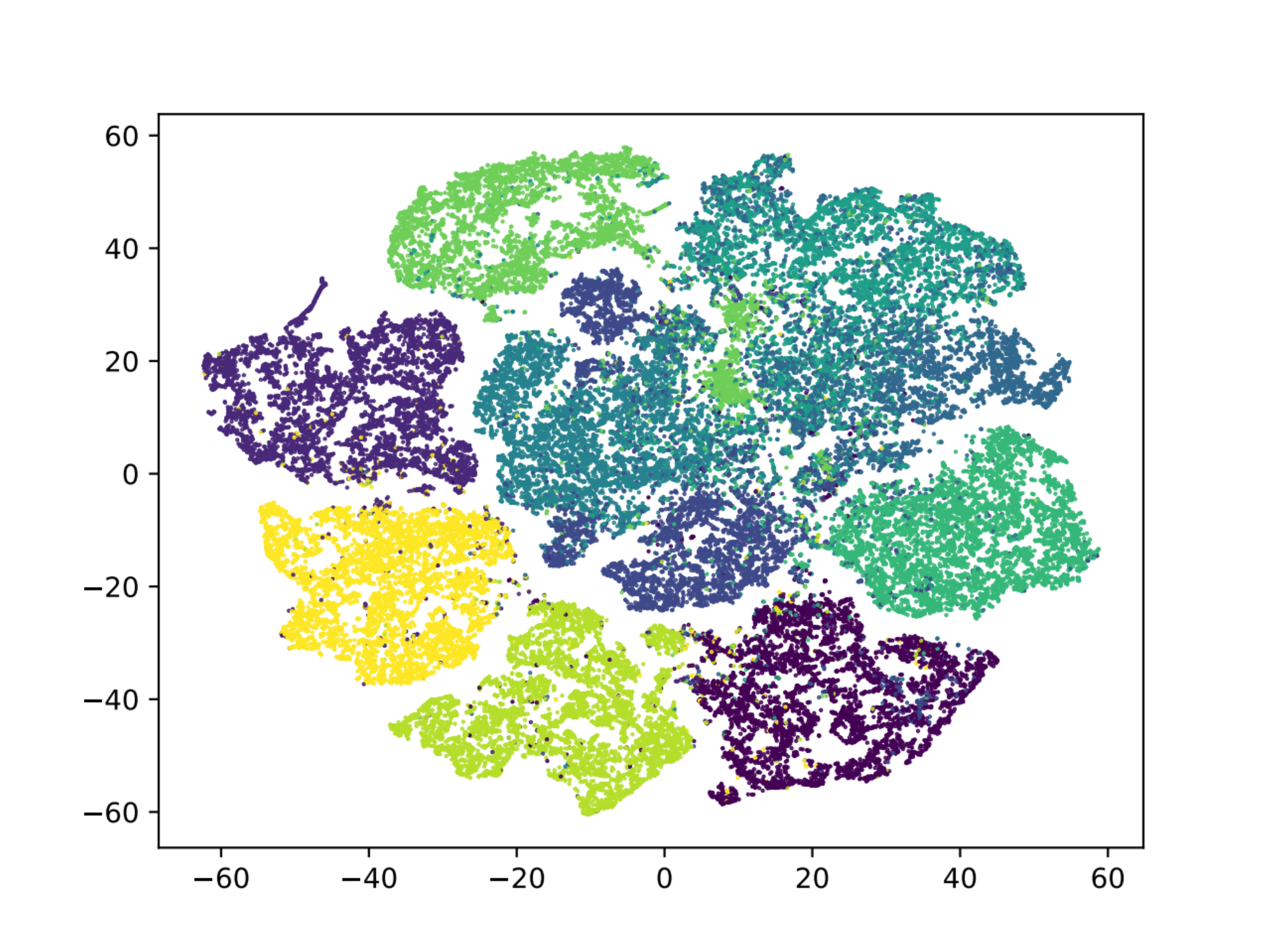}}
\subfigure[1000 epoch]{
\includegraphics[width=0.3\textwidth]{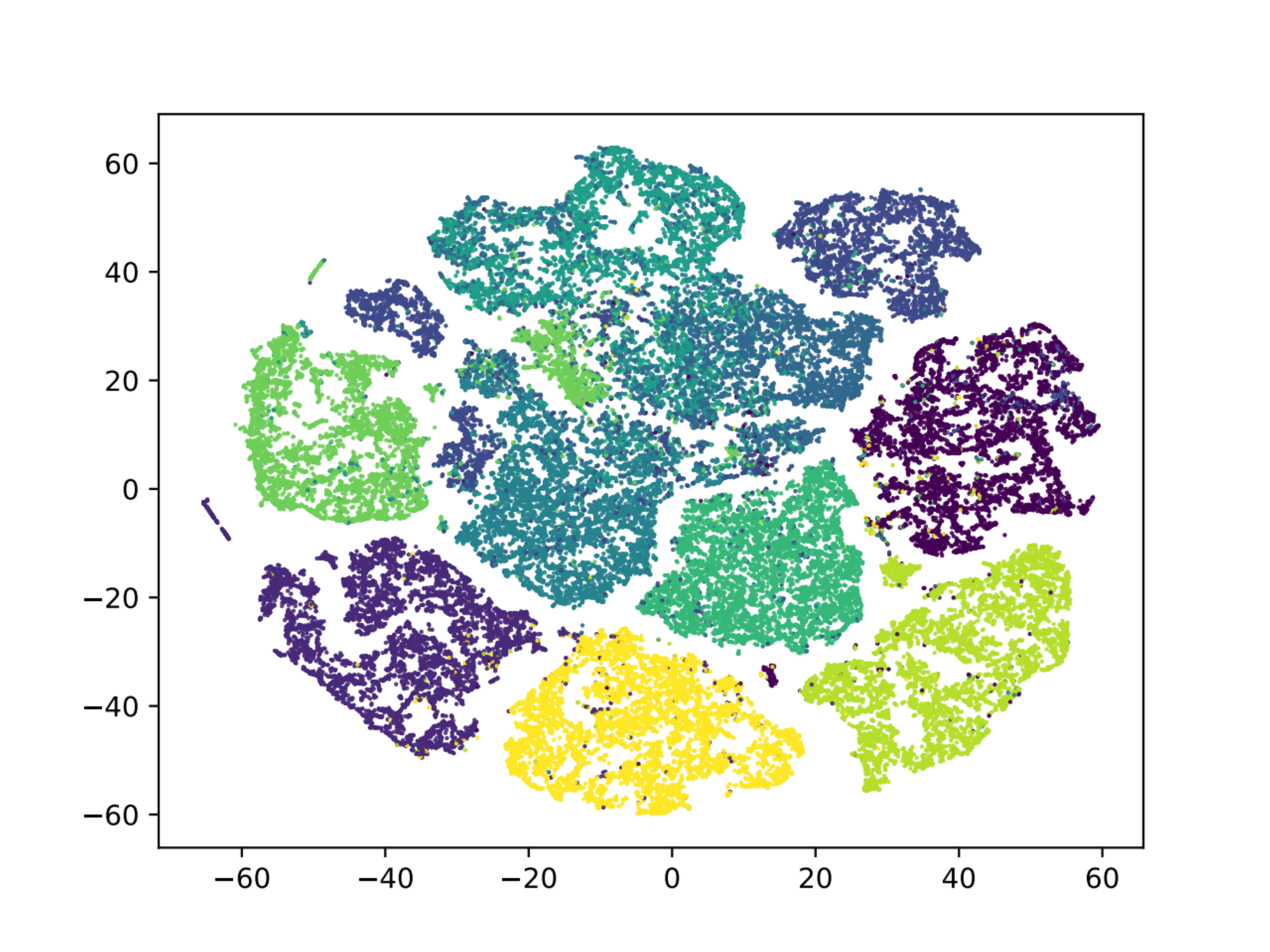}}
\caption{The t-SNE visualization of SACC on the CIFAR-10 dataset.}\vskip 0.1 in
\label{fig9}
\end{figure*}

In this section, we compare the proposed SACC method against both traditional and deep clustering methods on the five benchmark datasets.

The NMI, ACC, and ARI scores of different clustering methods are reported in Tables~\ref{table:comp_nmi}, \ref{table:comp_acc}, and \ref{table:comp_ari}, respectively. From Table~\ref{table:comp_nmi}, we can observe that the deep clustering methods can significantly outperform the traditional (non-deep) clustering methods, due to the representation learning ability of the deep neural networks. In terms of the proposed method, SACC achieves the best NMI score on the five benchmark datasets. Especially, on the CIFAR-10 and ImageNet-Dogs datasets, SACC achieves NMI(\%) scores of 76.5 and 45.5, respective, where the second best NMI(\%) scores are only 68.1 and 40.1, respectively. On the other three datasets, SACC also obtains better NMI scores than the other deep clustering methods. Similar advantages can be observed in Tables~\ref{table:comp_acc} and \ref{table:comp_ari}, where our SACC method also achieves the best ACC and ARI scores on all the five benchmark datasets.

Besides the quantitative evaluation, we further provide visual analysis on the clustering results of our SACC method. Specifically, Fig.~\ref{fig5} illustrates the confusion matrices between the true and predicted labels on the five image datasets.
As shown in Fig.~\ref{fig5},
clear block-diagonal structures can be observed in the confusion matrices for the CIFAR-10, STL-10, and ImageNet-10 datasets. Even for the more challenging datasets of CIFAR-100 and ImageNet-Dogs, we can still observe the block-diagonal structures, though they are not as clear as the other three datasets. Notably, even for the challenging datasets like CIFAR-100 and ImageNet-Dogs, our SACC method can still yield better clustering performance than the state-of-the-art deep clustering methods (as shown in Tables~\ref{table:comp_nmi}, \ref{table:comp_acc}, and \ref{table:comp_ari}).

\subsection{Ablation Study}

In this section, we conduct the ablation study on the CIFAR-10 dataset to test the influence of the strong and strong augmentations and that of the two contrastive projectors in our SACC method.

\begin{table}[]
\renewcommand\arraystretch{1.1}
\centering
\caption{The NMI(\%), ACC(\%), and ARI(\%) by SACC using different combinations of weak and strong augmentations.}
\label{tab3}
\begin{tabular}{p{4.6cm}p{0.8cm}<{\centering}p{0.8cm}<{\centering}p{0.8cm}<{\centering}}
\toprule
Augmentations    & NMI  & ACC  & ARI  \\ \midrule
Weak+Weak        & 68.1 & 76.6 & 60.6 \\
Weak+Strong      & 72.6 & 82.6 & 67.9 \\
Weak+Weak+Strong & \textbf{76.5} & \textbf{85.1} & \textbf{72.4} \\ \bottomrule
\end{tabular}
\end{table}

\begin{table}[]\vskip 0.01 in
\renewcommand\arraystretch{1.1}
\centering
\caption{The NMI(\%), ACC(\%), and ARI(\%) by SACC using one or two of its contrastive projectors.}
\label{tab4}
\begin{tabular}{p{4.6cm}p{0.8cm}<{\centering}p{0.8cm}<{\centering}p{0.8cm}<{\centering}}
\toprule
Used Projectors                             & NMI  & ACC  & ARI  \\ \midrule
With Only Cluster Projector              & 67.5 & 73.4 & 58.8 \\
With Only Instance Projector             & 69.2 & 75.5 & 61.5 \\
With Both Projectors                     & \textbf{76.5} & \textbf{85.1} & \textbf{72.4} \\
\bottomrule
\end{tabular}\vskip 0.08 in
\end{table}

\subsubsection{Influence of Strong and Weak Augmentations}

In our SACC framework, three augmentation views are utilized, including a strongly augmented views and two weakly augmented views (as shown in Fig.~\ref{fig2}). In the section, we test the influence of the three augmentation views. Note that at least two augmentation views should be preserved to make contrastive learning feasible. As shown in Table~\ref{tab3}, using a weak augmentation and a strong augmentation leads to better clustering performance than using two weak augmentations. Further, the proposed SACC method using all three augmentations significantly outperforms that variant of using two augmentations (weak+weak or weak+strong), which demonstrate the substantial benefits brought in by  our network architecture with three augmentation views.

\subsubsection{Influence of Two Contrastive Projectors}

In SACC, we utilize two projectors, namely, the instance projector and the cluster projector, for the instance-level and cluster-level contrastive learning, respectively. In the section, we test the influence of these two projectors by evaluating the performance of SACC with one of the two projectors removed. Note that in SACC, the clustering result is obtained in the cluster projector. For the variant with the cluster projector removed, we obtain the clustering result by performing $K$-means on the feature representation learned by the instance projector. As shown in Table~\ref{tab4}, using both projectors can lead to significantly better clustering performance (w.r.t. NMI, ACC, and ARI) than using only one of the two projectors, which confirm the advantage of the joint instance-level and cluster-level contrastive learning in our SACC method.

\subsection{Convergence Analysis}

In this section, we evaluate the convergence of the proposed SACC method. Specifically, the NMI, ACC, and ARI scores of the proposed method are recorded for each 100 epochs, which are then plotted in  Fig.~\ref{fig6}, ~\ref{fig7} and ~\ref{fig8}. As shown Fig.~\ref{fig6}, ~\ref{fig7} and ~\ref{fig8}, the NMI, ACC, and ARI scores of SACC consistently increase as the number of epochs grows. On most of the datasets, the proposed SACC method can reach high-quality clustering results when the number of epochs is greater than 500. In this paper, the number  of epochs is set to 1,000 for all datasets.

Further, we visualize convergence of SACC  by performing t-SNE \cite{van2008visualizing} on the learned feature representations the learned feature representation on the CIFAR-10 dataset, which is shown in Fig.~\ref{fig9}, where different colors denote different ground-truth labels. As can be observed in Fig.~\ref{fig9}, at the beginning, the data samples are mostly mixed. After training 100 epochs, many samples in the same class have been grouped closer. As the learning process proceeds, the distribution of the data samples reach stability after about 500 epochs, where the separability of the samples in most classes becomes relatively clear.

\section{Conclusion}\label{sec5}

In this paper, we propose a novel deep clustering approach termed SACC. Different from the previous contrastive learning based deep clustering approaches which typically use some weak augmentations with limited transformations, our SACC approach is able to jointly leverage strong and weak augmentations for enhancing the simultaneous contrastive representation learning and clustering. In particular, we utilize a backbone network with triply-shared weights for three augmentation views, including a strongly augmented view and two weakly augmented views. Three views of representations (for weakly and strongly augmented samples) can be obtained from the backbone network, which are then fed to two types of projectors, namely, the instance projector and the cluster projector, so as to enable the instance-level contrastive learning and the cluster-level contrastive learning, respectively. Further, the unsupervised training is performed to simultaneously optimize the backbone and the two projectors in an end-to-end manner, where the final clustering result can therefore be achieved in the cluster projector. Extensive experiments are conducted on five benchmark image datasets, which have confirmed the superior clustering performance of the proposed SACC approach over the state-of-the-art deep clustering approaches.

\bibliographystyle{IEEEtran}
\bibliography{refs}

\begin{IEEEbiography}[{\includegraphics[width=1in,height=1.25in,clip,keepaspectratio]{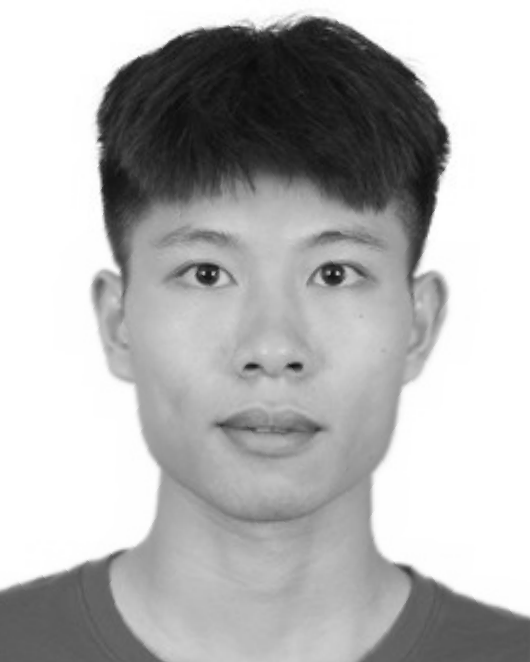}}]{Xiaozhi Deng}
	received the B.S. degree in 2020 from South China Agricultural University, Guangzhou, China, where he is currently pursuing the master degree in computer science with the College of Mathematics and Informatics. His research interests include clustering analysis and deep learning.
\end{IEEEbiography}

\begin{IEEEbiography}[{\includegraphics[width=1in,height=1.25in,clip,keepaspectratio]{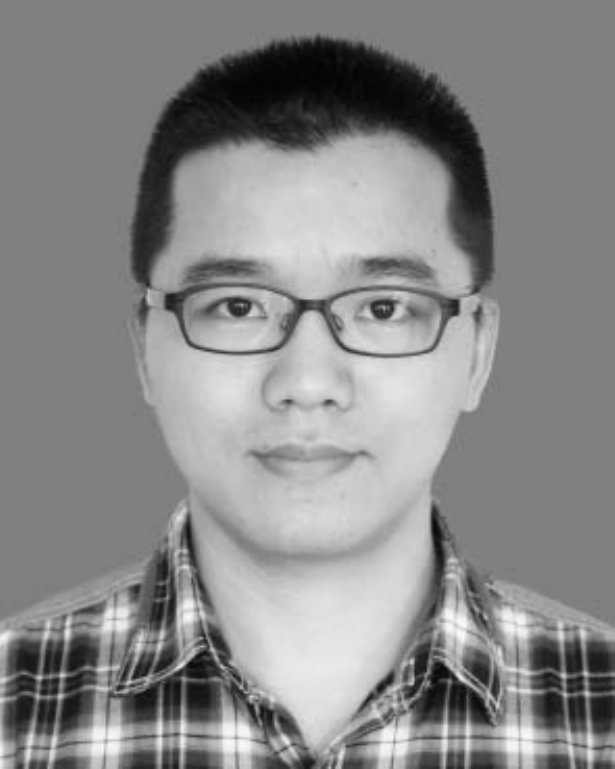}}]{Dong Huang}
	received the B.S. degree in computer science in 2009 from South China University of Technology, Guangzhou, China. He received the M.Sc. degree in computer science in 2011 and the Ph.D. degree in computer science in 2015, both from Sun Yat-sen University, Guangzhou, China. He joined South China Agricultural University in 2015, where he is currently an Associate Professor with the College of Mathematics and Informatics. From July 2017 to July 2018, he was a visiting fellow with the School of Computer Science and Engineering, Nanyang Technological University, Singapore. His research interests include data mining and machine learning. He has published more than 50 papers in international journals and conferences, such as IEEE TKDE, IEEE TCYB, IEEE TSMC-S, ACM TKDD, SIGKDD, AAAI, and ICDM. He was the recipient of the 2020 ACM Guangzhou Rising Star Award.
\end{IEEEbiography}

\begin{IEEEbiography}[{\includegraphics[width=1in,height=1.25in,clip,keepaspectratio]{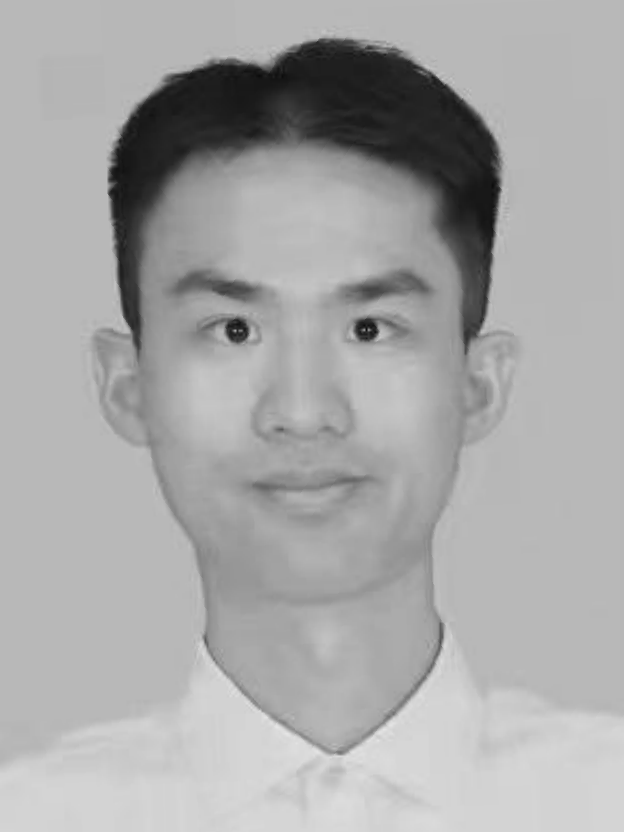}}]{Ding-Hua Chen}
	received the B.S. degree in computer science and technology in 2019 from South China Agricultural University, Guangzhou, China, where he is currently pursuing the master degree in computer science with the College of Mathematics and Informatics. His current research interests include deep learning and clustering analysis.
\end{IEEEbiography}

\begin{IEEEbiography}[{\includegraphics[width=1in,height=1.25in,clip,keepaspectratio]{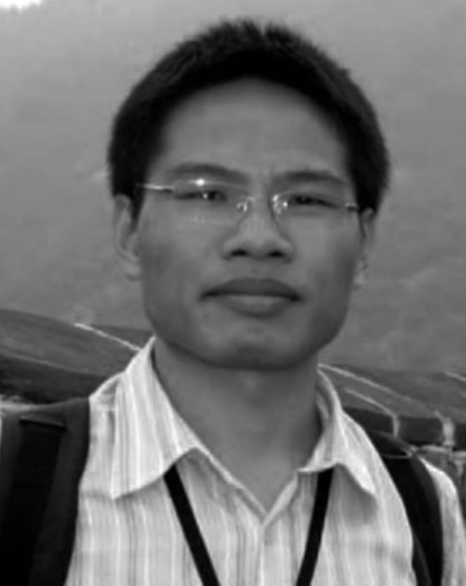}}]{Chang-Dong Wang}
	received the B.S. degree in applied mathematics in 2008, the M.Sc. degree in computer science in 2010, and the Ph.D. degree in computer science in 2013, all from Sun Yat-sen University, Guangzhou, China. He was a visiting student at the University of Illinois at Chicago from January 2012 to November 2012. He is currently an Associate Professor with the School of Data and Computer Science, Sun Yat-sen University, Guangzhou, China. His current research interests include machine learning and data mining. He has published more than 100 scientific papers in international journals and conferences such as IEEE TPAMI, IEEE TKDE, IEEE TNNLS, IEEE TSMC-C, ACM TKDD, Pattern Recognition, SIGKDD, ICDM and SDM. His ICDM 2010 paper won the Honorable Mention for Best Research Paper Award. He was awarded 2015 Chinese Association for Artificial Intelligence (CAAI) Outstanding Dissertation.
\end{IEEEbiography}

\begin{IEEEbiography}[{\includegraphics[width=1in,height=1.25in,clip,keepaspectratio]{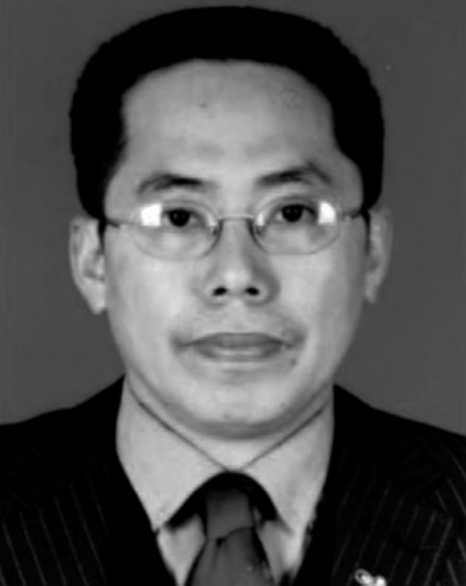}}]{Jian-Huang Lai}
	received the M.Sc. degree in applied mathematics in 1989 and the Ph.D. degree in mathematics in 1999 from Sun Yat-sen University, China. He joined Sun Yat-sen University in 1989 as an Assistant Professor, where he is currently a Professor with the School of Data and Computer Science. His current research interests include the areas of digital image processing, pattern recognition, multimedia communication, wavelet and its applications. He has published more than 200 scientific papers in the international journals and conferences on image processing and pattern recognition, such as IEEE TPAMI, IEEE TKDE, IEEE TNN, IEEE TIP, IEEE TSMC-B, Pattern Recognition, ICCV, CVPR, IJCAI, ICDM and SDM. Prof. Lai serves as a Standing Member of the Image and Graphics Association of China, and also serves as a Standing Director of the Image and Graphics Association of Guangdong.
\end{IEEEbiography}

\end{document}